%% file: main.tex
\documentclass[10pt,twocolumn,letterpaper]{article}

\usepackage{cvpr}              

\usepackage{setspace}

\usepackage{dsfont}

\input{preamble}

\include{def}

\definecolor{cvprblue}{rgb}{0.21,0.49,0.74}
\usepackage[pagebackref,breaklinks,colorlinks,citecolor=cvprblue]{hyperref}

\def\paperID{7276} 
\def\confName{CVPR}
\def\confYear{2025}

\def\mytitle{MatIR: A Hybrid Mamba-Transformer Image Restoration Model
}
\title{\mytitle}

\author{Juan Wen$^{1,2}$,\enspace Weiyan Hou$^{1}$,\enspace Luc Van Gool$^{2,3,4}$, \enspace Radu Timofte$^{5}$\\
\textsuperscript{1}Zhengzhou University,\enspace
\textsuperscript{2}ETH Z\"{u}rich,\enspace
\textsuperscript{3}KU Leuven,\enspace 
\textsuperscript{4}INSAIT, Sofia University,\enspace
\textsuperscript{5}University of Wurzburg\\
{\small \textbf{\url{https://github.com/wenjuan7275/MatIR}}}
}

\begin{document}
\maketitle
\input{sec/0_abstract}    
\input{sec/1_intro}

\input{sec/2_related_work}

\input{sec/3_method}
\input{sec/4_exp}

\input{sec/5_conclusion}

{
    \small
    \bibliographystyle{ieeenat_fullname}
    \bibliography{main}
}

\end{document}

%% file: preamble.tex
\usepackage[dvipsnames]{xcolor}


%% file: def.tex
\usepackage{url}            
\usepackage{xcolor}         

\usepackage{times}
\usepackage{microtype} 
\usepackage{graphicx} 
\usepackage{wrapfig}
\usepackage{balance} 

\usepackage{algorithm,algpseudocode}

\usepackage{helvet}
\usepackage{courier}

\usepackage{makecell}
\usepackage{graphicx}
\usepackage{color}
\usepackage{amsfonts}
\usepackage{amsmath}
\usepackage{amssymb}

\usepackage{bm}

\usepackage{multirow}
\usepackage{multicol} 
\usepackage{booktabs}

\usepackage{colortbl} 
\usepackage{caption} 
\usepackage[dvipsnames]{xcolor} 
\colorlet{colorFst}{Green!25}       
\colorlet{colorSnd}{SpringGreen!45} 
\colorlet{colorTrd}{Yellow!30}      
\colorlet{colorLow}{darkgray!30}    
\colorlet{colorDeg}{Orange!30}      
\newcommand{\fs}{\cellcolor{colorFst}\bf}   
\newcommand{\nd}{\cellcolor{colorSnd}}      
\newcommand{\rd}{\cellcolor{colorTrd}}      

\DeclareMathAlphabet\mathbfcal{OMS}{cmsy}{b}{n}

\def\0{{\bf 0}}
\def\1{{\bf 1}}

\usepackage{ntheorem}

\newtheorem*{*thm}{Theorem}

\newtheorem*{*lemma}{Lemma}

\usepackage{stackengine}
\usepackage{tabularx}
\usepackage{threeparttable}
\usepackage{arydshln}

\definecolor{mygray}{gray}{.9}

\usepackage{pifont}

\usepackage{adjustbox}

\newlength \g

\usepackage{enumitem}

%% file: sec/0_abstract.tex
\begin{abstract}

In recent years, Transformers-based models have made significant progress in the field of image restoration by leveraging their inherent ability to capture complex contextual features. Recently, Mamba models have made a splash in the field of computer vision due to their ability to handle long-range dependencies and their significant computational efficiency compared to Transformers. However, Mamba currently lags behind Transformers in contextual learning capabilities. To overcome the limitations of these two models, we propose a Mamba-Transformer hybrid image restoration model called MatIR. Specifically, MatIR cross-cycles the blocks of the Transformer layer and the Mamba layer to extract features, thereby taking full advantage of the advantages of the two architectures. In the Mamba module, we introduce the Image Inpainting State Space (IRSS) module, which traverses along four scan paths to achieve efficient processing of long sequence data. In the Transformer module, we combine triangular window-based local attention with channel-based global attention to effectively activate the attention mechanism over a wider range of image pixels. Extensive experimental results and ablation studies demonstrate the effectiveness of our approach.
\end{abstract}

%% file: sec/1_intro.tex
\section{Introduction}
\label{sec:intro}
Image restoration aims to recover a clear and high-quality image from degraded or corrupted input. This is a long-standing problem in computer vision and encompasses a wide range of sub-problems such as super-resolution, image denoising, and deblurring. With the introduction of modern deep learning models such as CNNs~\cite{DnCNN,dong2014learning,lim2017enhanced,zhang2018residual,dai2019second} and Transformers~\cite{chen2021pre,liang2021swinir,chen2023activating,li2023grl,chen2023dual,wen2023super,wen2024empowering}, the state-of-the-art performance has been continuously improved over the past few years. Tasks such as denoising, deblurring, and super-resolution require models that can accurately reconstruct image details while preserving structural information. Traditional convolutional-based methods are often unable to capture long-range dependencies that are critical for tasks involving large or severely degraded images. Recent advances in deep learning, such as the Transformer architecture, have shown promise in capturing global dependencies in images. However, the computational cost of Transformers grows quadratically with the sequence length, limiting their scalability, especially for high-resolution image restoration tasks, which offer a global receptive field at the expense of quadratic complexity.

\begin{figure}[t]
  \hspace{-10pt}
   \includegraphics[width=1.08\linewidth]{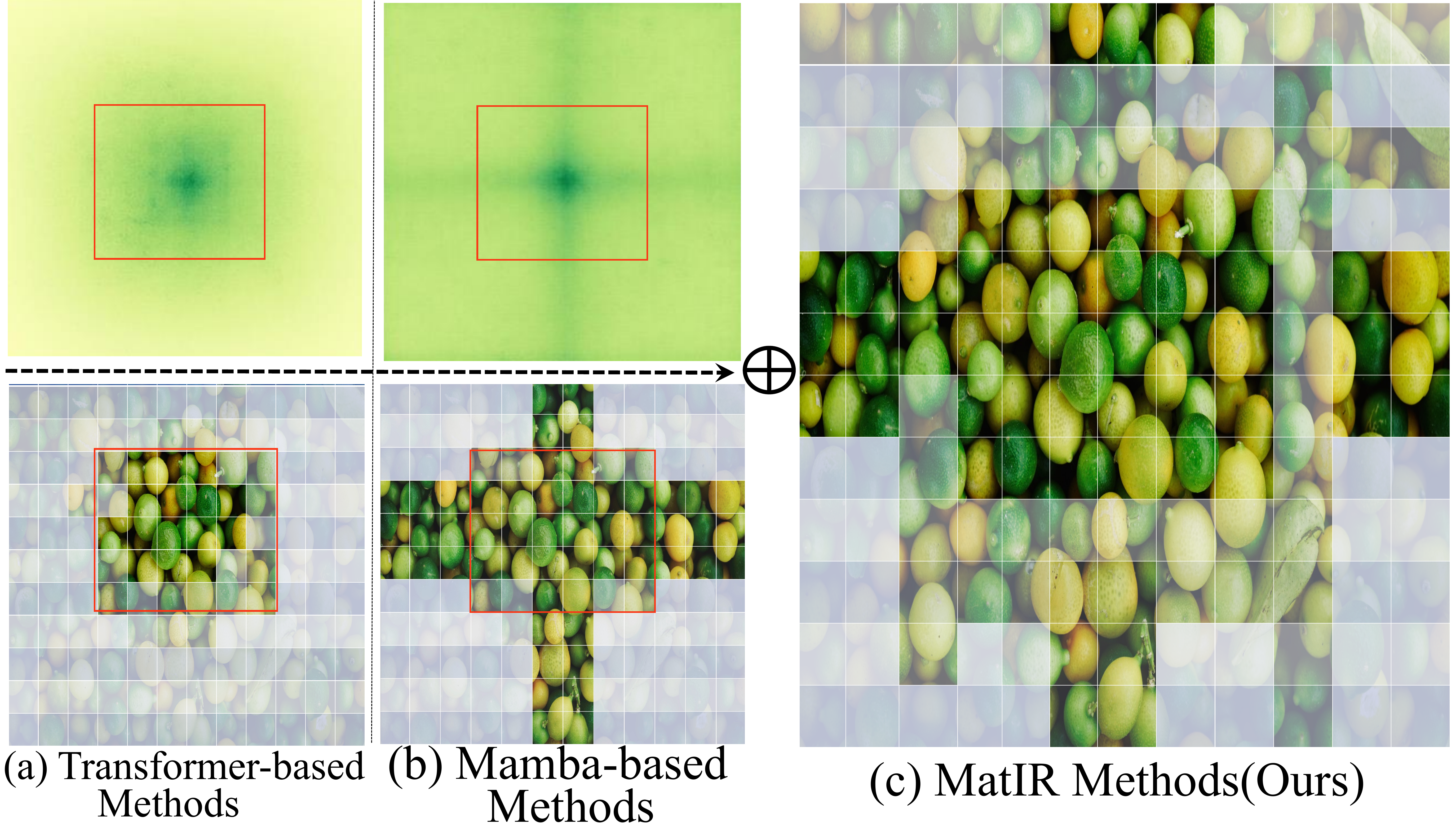}
   \vspace{-7mm}
   \caption{(a) Visualization of the effective receptive field (ERF) of the Transformer basic model~\cite{luo2016understanding,ding2022scaling}. The figure below shows the advantage of being larger than Manba in the context neighbor receptive field. (b) Visualization of the ERF of the Manba basic model ~\cite{liu2024vmamba}. The figure below shows the advantage of being larger than Transformer in long sequence linear receptive fields. (c) Our proposed MatIR: A Hybrid Mamba-Transformer Image Restoration Model achieves a more significant effective receptive field.}
   \vspace{-5mm}
   \label{fig:comp_ir}
\end{figure}

Recently Mamba has become increasingly prominent in the field of computer vision due to its ability to manage long-distance dependencies and its significant computational efficiency relative to Transformers. The Mamba architecture is a new sequence model that achieves efficient processing of long sequence data by introducing the concept of a state space model (SSM). State-space models (SSM) leverage state-space representations to achieve linear computational complexity and can efficiently process long sequences without compromising accuracy. With linear computational complexity related to sequence length, it shows efficiency and effectiveness in handling long-distance dependencies in sequence modeling tasks. However, existing research shows that Mamba lags behind Transformers in contextual learning (ICL) capabilities.~\cite{mamba,mamba2,waleffe2024empirical} Draw inspiration from recent advances in the development of modern deep neural networks.
In this work, our aim is to compensate for the shortcomings of the models while leveraging their respective strengths. We propose a hybrid Mamba-Transformer image restoration model, called MatIR. This is a novel hybrid architecture that leverages the strengths of the Mamba architecture (known for its memory efficiency in processing long sequences) and the Transformer (excelling in contextual learning and information retrieval)~\cite{mamba,mamba2,waleffe2024empirical}. By combining these two approaches, MatIR aims to provide a powerful and efficient solution for a variety of image restoration tasks~\cite{ray2024cfat,liu2024vmamba}.

Specifically, 1) the shallow feature extraction stage uses simple convolutional layers to extract shallow features. Then 2) the deep learning feature extraction stage uses a Transformer stacked with Mamba layers. In the Transformer layer, we use triangular window local attention (TWLA) and channel global attention (CGA) to effectively activate the attention mechanism over a wider range of image pixels to improve the performance of this module. In the Mamba layer, we use the Image Restoration State Space (IRSS) module, which traverses along four scanning paths to achieve efficient processing of long sequence data from different directions and paths. The performance and throughput are improved while maintaining a manageable memory footprint. As the three core components of our MatIR, TWLA and CGA each activate more input pixels from the range of local and global, triangular window and rectangular window in attention to achieve higher quality image restoration. IRSS creates linear computational complexity information related to sequence length in the state space from four different directional paths: left, top, right, and bottom, showing efficiency and effectiveness in handling long-distance dependencies in sequence modeling tasks. Finally, 3) the high-quality image reconstruction stage aggregates shallow and deep features to produce high-quality output images. With local and global effective receptive fields as well as efficient memory management and computation, our MatIR becomes a new alternative to the image restoration backbone.

In short, our main contributions can be summarized as follows:
\begin{itemize}
	\item We apply the state-space model and Transformer attention mechanism to the field of image restoration through extensive experiments, thus formulating MatIR, which greatly improves the computational efficiency while maintaining performance.
    \item We propose the Image Restoration State Space (IRSS) module, which scans and traverses along four different paths to achieve efficient processing of long sequence data.
    \item We propose a Triangular Window Local Attention (TWLA) block and a Channel Global Attention (CGA) block each to attentionally activate more input pixels from the range of local and global, triangular window and rectangular window respectively in attention to achieve higher quality image restoration.
	\item Through comprehensive evaluation on multiple benchmark datasets, our method has superior performance compared with other state-of-the-art methods, providing a powerful and promising backbone solution for image restoration.
\end{itemize}

%% file: sec/2_related_work.tex
\begin{figure*}[t]
  \centering
   \vspace{-3mm}
   \includegraphics[width=0.75\linewidth]{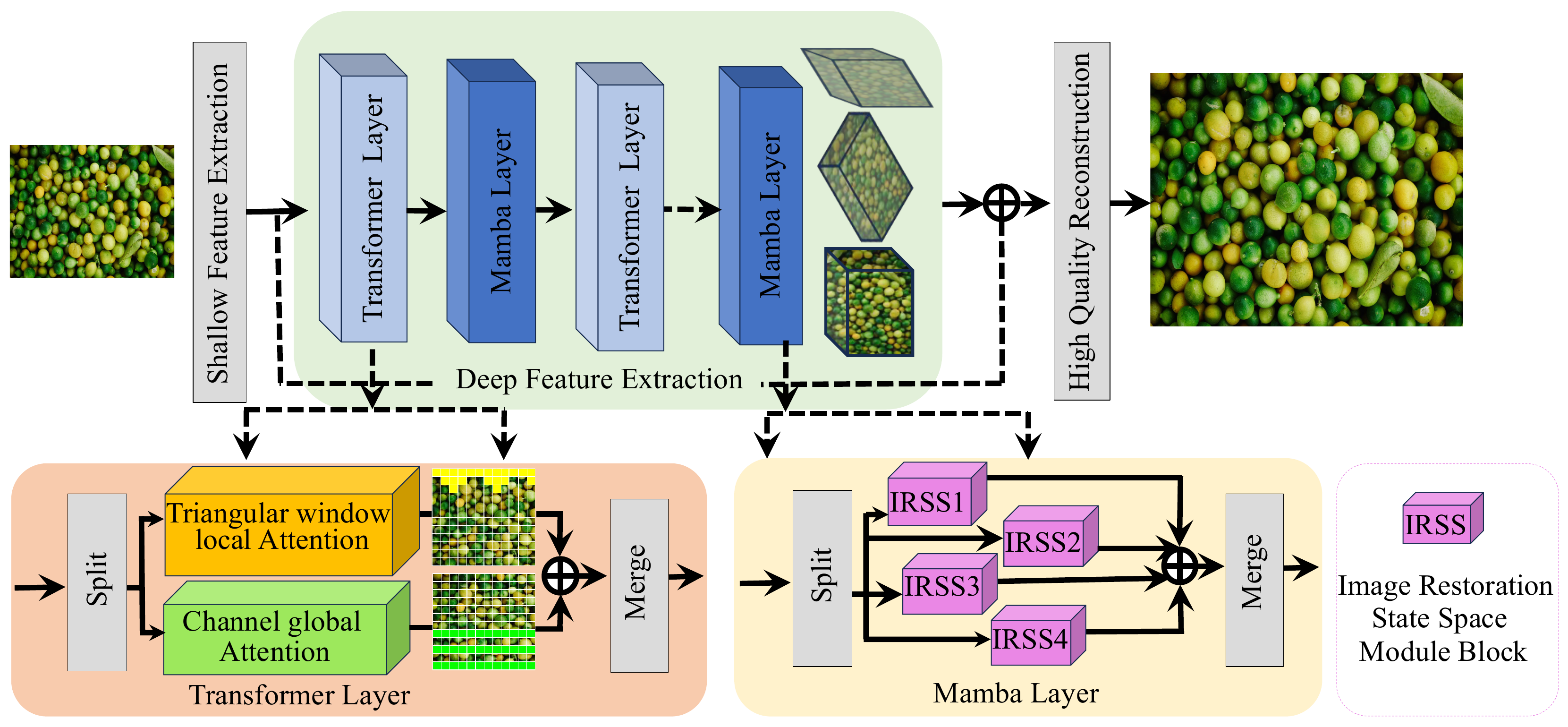}
   \vspace{-4mm}
   \caption{Overall network architecture of our MatIR.}
   \vspace{-4mm}
   \label{fig:MatIR1}
\end{figure*}


\vspace{-1mm}
\section{Related Work}
\paragraph{Image Restoration.}
Image restoration is a long-standing problem in computer vision. In the past decade, many efforts have been made in various fields to improve the performance of deep learning methods, including image restoration. Pioneered by SRCNN ~\cite{dong2014learning}, deep learning was introduced to image restoration super-resolution through a simple three-layer convolutional neural network (CNN). Since then, many studies have explored various architectural enhancements to improve the performance ~\cite{Kim_2016_vdsr, lim2017edsr, Zhang_2018_rdn, zhang2018rcan, Dai_2020_san, Niu_2020_han, Mei_2021_nlsa, mei2020image, kim2016deeply}. VDSR ~\cite{Kim_2016_vdsr} implemented a deeper network, and DRCN~\cite{kim2016deeply}proposed a recursive structure. EDSR~\cite{lim2017edsr} and RDN~\cite{Zhang_2018_rdn} developed new residual blocks to further improve the ability of CNN in SR. However, despite the success of CNN, the receptive field of CNN is inherently limited, making it difficult to capture long-range dependencies. 

In recent years, Vision Transformer (ViT)~\cite{dosovitskiy2020image} and its variants~\cite{liu2021swin, Chu_2021_twin, Wang_2022_pvt} introduced the self-attention mechanism into image processing, allowing the model to learn global relationships. Based on this, IPT~\cite{chen2021pre}successfully attempted to utilize transformer-based networks for various image restoration tasks. Since then, multiple techniques have been developed to enhance the performance of the image restoration Transformer. These include the shifted window self-attention implemented by SwinIR~\cite{liang2021swinir} and CAT~\cite{chen2022cross}, the grouped multiscale self-attention mechanism of ELAN~\cite{zhang2022efficient}, the sparse attention of ART~\cite{zhang2023accurate} and OmniSR~\cite{omni_sr}, and the anchored self-attention mechanism of GRL~\cite{li2023grl}, the multi-attention mechanism DART, and DISR~\cite{wen2023super,wen2024empowering} all of which aim to expand the range of the receptive field to achieve better results. However, it provides a global receptive field at the cost of quadratic complexity. The quadratic computational complexity of self-attention in terms of sequence length poses a challenge, especially for high-resolution images.

\vspace{-5mm}
\paragraph{State Space Models.}
Recently, Mamba has become increasingly prominent in the field of computer vision due to its ability to manage long-distance dependencies and its significant computational efficiency relative to Transformers. The Mamba architecture is a new sequence model that achieves efficient processing of long sequence data by introducing the concept of a state space model (SSM)~\cite{mamba,mamba2,waleffe2024empirical}. State-space models (SSM) leverage state-space representations to achieve linear computational complexity and can efficiently process long sequences without compromising accuracy. With linear computational complexity related to sequence length, it shows efficiency and effectiveness in handling long-distance dependencies in sequence modeling tasks. However, existing research shows that Mamba lags behind Transformers in contextual learning (ICL) capabilities~\cite{waleffe2024empirical}. The trade-off dilemma between efficient computation and global modeling has not been essentially resolved~\cite{ray2024cfat,liu2024vmamba,guo2025mambair}. 
Considering the limitations of two current state-of-the-art models, we explore the potential of the hybrid Mamba-Transformer approach for image inpainting. In this paper, based on the effectiveness of the Mamba and Transformer models, we propose three core components, the Image Restoration State Space (IRSS) module, which scans and traverses along four different paths to achieve efficient processing of long-sequence data. The Triangular Window Local Attention (TWLA) block and the Channel Global Attention (CGA) block, each attentionally activate more input pixels from the range of local and global, triangular and rectangular windows to achieve higher quality image restoration.

%% file: sec/3_method.tex
\vspace{-1mm}
\section{Methodology}

\subsection{Preliminaries}
\label{sec:SSM_Module}
\vspace{-2mm}
Structured state space sequence models (S4) are a recent class of sequence models for deep learning that are broadly related to RNNs, CNNs,
and classical state space models.
They are inspired by a particular continuous system 
that maps a 1-dimensional function or sequence $x(t) \in \mathds{R} \mapsto y(t) \in \mathds{R}$ through an implicit latent state \( h(t) \in \mathds{R}^N \). %

Formally, this system can be formulated as a linear ordinary differential equation (ODE)~\cite{mamba,mamba2,waleffe2024empirical}:
\vspace{-3mm}
\begin{equation}
\begin{aligned}
\label{eq:ssm}
    h'(t)&={\rm \textbf{A}}h(t)+{\rm \textbf{B}}x(t),\\
    y(t)&={\rm \textbf{C}}h(t)+{\rm \textbf{D}}x(t),
\end{aligned}
\vspace{-3mm}
\end{equation}

\noindent
where $N$ is the state size, ${\rm \textbf{A}} \in \mathds{R}^{N\times N}$, ${\rm \textbf{B}} \in \mathds{R}^{N \times 1}$, ${\rm \textbf{C}} \in \mathds{R}^{1\times N}$, and ${\rm \textbf{D}} \in \mathds{R}$.

After that, the discretization process is typically adopted to integrate \cref{eq:ssm} into practical deep learning algorithms. Specifically, denote $\rm \Delta$ as the timescale parameter to transform the continuous parameters ${{\rm \textbf{A}}}$, ${{\rm \textbf{B}}}$ to discrete parameters $\overline{{\rm \textbf{A}}}$, $\overline{{\rm \textbf{B}}}$. The commonly used method for discretization is the zero-order hold rule (ZOH), which is defined as follows:
\vspace{-2mm}
\begin{equation}
\begin{aligned}
    \overline{\rm \textbf{A}} &= {\rm exp}({\rm {\Delta \textbf{A}}}),\\
    \overline{\rm \textbf{B}}&=({\rm {\Delta \textbf{A}}})^{-1}({\rm exp(\textbf{A})}-\textbf{I})\cdot {\rm \Delta \textbf{B}}.
\end{aligned}
\vspace{-2mm}
\end{equation}

After the discretization, the discretized version of \cref{eq:ssm} with step size $\rm \Delta$ can be rewritten in the following RNN form:

\vspace{-2mm}
\begin{equation}
\begin{aligned}
\label{eq:discret-ssm}
    h_k&=\overline{\rm \textbf{A}}h_{k-1}+\overline{\rm \textbf{B}}x_k,\\
    y_k&={\rm \textbf{C}}h_k+{\rm \textbf{D}}x_k.
\end{aligned}
\vspace{-2mm}
\end{equation}

Furthermore,~\cref{eq:discret-ssm} can also be mathematically equivalently transformed into the following CNN form:
\vspace{-2mm}
\begin{equation}
\begin{aligned}
\label{eq:cnn-form}
\overline{\rm \textbf{K}}&\triangleq(\mathrm{\textbf{C}} \overline{\mathrm{\textbf{B}}},{\rm{\textbf{C}}}\overline{\rm \textbf{A}}\overline{\rm \textbf{B}},\cdots,{\rm{\textbf{C}}}{\overline{\rm \textbf{A}}}^{L-1}\overline{\rm \textbf{B}}),\\
{\rm \textbf{y}}&={\rm \textbf{x}} \ast \bm {\overline{\rm \textbf{K}}},
\end{aligned}
\vspace{-2mm}
\end{equation}

\noindent
where $L$ is the length of the input sequence, $\ast$ denotes the convolution operation, and $\overline{\rm \textbf{K}} \in \mathds{R}^L$ is a structured convolution kernel.

Commonly, the model uses the convolutional mode \cref{eq:cnn-form} for efficient parallelizable training (where the whole input sequence
is seen ahead of time) and switched into recurrent mode~\cref{eq:discret-ssm} for efficient autoregressive inference (where the inputs are
seen one timestep at a time). An important property that we can see from the equation is that the dynamics of the model remain constant over time. This property is called linear time invariance. From this attribute, we can see the advantage of its model dynamics with Transformer, Transformer which increases quadratically with the computational cost and sequence length. The state-space model (SSM) utilizes state-space representation to achieve linear computational complexity, which has linear computational complexity related to the sequence length and can efficiently process long sequences without affecting accuracy.

 \vspace{-1mm}
\subsection{Channel Global Attention Block}
\vspace{-1mm}
Recent block designs in Transformer-based restoration networks, such as those used in SwinIR, GRL, DART, and DISR~\cite{liang2021swinir,wen2023super,li2023efficient,wen2024empowering}, have demonstrated that channel-based global attention (CGA) performs exceptionally well in the field of image restoration. This suggests that applying CGA to a MatIR-based restoration network, by tailoring a novel block structure, holds great potential. So we input the data, Let the input feature map be $\mathbf{X}\in \mathds{R}^{C \times H \times W}$, where ${C}$ is the number of channels, ${H}$, ${W}$ are the spatial dimensions of the feature map. Flatten the input ${X}$ along the channel dimension to obtain $\mathbf{X}_{\text{flat}} \in\mathds{R}^{C\times N},\text{ where }N=H\times W$. Perform global pooling over the spatial dimensions to aggregate spatial information, obtaining a global representation for each channel:
$\mathbf{z}=\frac{1}{N}\sum_{i=1}^{N}\mathbf{X}_{\text{flat}}[:, i]$, where $\mathbf{z}\in \mathds{R}^C$ represents the global description for each channel. The attention mechanism directs computational resources toward the most information-rich parts of the input signal, enhancing model efficiency. The Transformer attention mechanism is based on Query$\mathbf{(Q)}$, Key$\mathbf{(K)}$ and Value$\mathbf{(V)}$
:
\vspace{-3mm}
\begin{equation}
\begin{aligned}
\label{eq:CGA}
    \text{Attention}(\mathbf{Q}, \mathbf{K}, \mathbf{V}) = \text{softmax}\left(\frac{\mathbf{Q}\mathbf{K}^\top}{\sqrt{d_k}}\right)\mathbf{V}.
\end{aligned}
\vspace{-2mm}
\end{equation}
In channel attention, we focus on the channel dimension and define:Query: $\mathbf{Q} = \mathbf{W}_q \mathbf{z},$
Key: $\mathbf{K} = \mathbf{W}_k \mathbf{z},$
Value: $\mathbf{V} = \mathbf{W}_v \mathbf{z},$
where $\mathbf{W}_q, \mathbf{W}_k, \mathbf{W}_v \in \mathds{R}^{C \times C}$ are learnable weight matrices. Specifically, the channel-based global attention mechanism operates across feature dimensions, allowing us to understand which features the model relies on when making a specific decision. By analyzing the attention weights across channels, we can gain insight into the model's reasoning process, which can help identify biases or areas for improvement. So we calculate the channel attention matrix 
\vspace{-4mm}
\begin{equation}
\begin{aligned}
\label{eq:CGA2}
\mathbf{A} = \text{softmax}\left(\frac{\mathbf{Q} \mathbf{K}^\top}{\sqrt{C}}\right),
\end{aligned}
\vspace{-4mm}
\end{equation}
where $\mathbf{A} \in \mathds{R}^{C \times C}$ is the attention weight matrix between channels. The output is obtained by weighting the channel representations:
\( \mathbf{z}_{\text{out}}=\mathbf{A}\mathbf{V}.\)
Reapply the channel attention result to the original input feature $\mathbf{X}$, usually through dot product or channel-by-channel weighting:
\( \mathbf{X}_{\text{CGA}}[:, i] = \mathbf{z}_{\text{out}} \odot \mathbf{X}_{\text{flat}}[:, i] \)
 where \( \odot \) denotes element-wise multiplication. The final Transformer-based channel global attention formula can be simplified to \( \mathbf{X}_{\text{CGA}} = \mathbf{X}\odot \text{Attention}(\mathbf{W}_q \mathbf{z}, \mathbf{W}_k \mathbf{z}, \mathbf{W}_v \mathbf{z}) \). This mechanism captures the global dependencies between channels through self-attention, which helps enhance the model's understanding of multi-channel features. However, when CGA extracts features through a rectangular window, the model may experience boundary distortion and encounter limitations in the shifted configuration. Addressing these challenges can further improve the effectiveness of CGA in repairing networks.
\subsection{Triangular Window Local Attention Block}
In computer vision, the properties of a pixel depend on itself and its neighbors. Therefore, during self-attention, the edge pixels of the rectangular window are not explored as effectively as the internal pixels, resulting in boundary distortion. When implemented separately, the above problems are obvious in both rectangular and triangular windows ~\cite{ray2024cfat}. Therefore, to solve the limitation of CGA introduced by MatIR, we adopted a triangular attention window in MatIR to extract features in a local range. Assume the input feature is $\mathbf{X} \in \mathds{R}^{N \times D}$, Where: $N$ is the number of input features, $D$ is the dimension of the feature. We also assume that each point $i$ has $k$ neighbors in its local area, forming a triangular structure. For each point $i$, select its local $k$ neighbors to form a local area. The triangular local structure contains the following information: Center Point $X_{\text{i}}$, Neighborhood set $
\left\{X_j\right\}_{j \in \mathcal{N}(i)}
$, The geometric relationships that make up a triangle, such as edge vectors and angle information: Edge Vector $e_{i j}=X_j-X_i$, Angle information: Determined by the relationship between neighbors, for example, $
\theta_{i j k}$ represents the angle between $j,k$. In order to utilize the triangle geometric structure information, the edge feature map is designed $f_{i j}=\phi\left(e_{i j}\right)$ Where $\theta$ is the multi-layer perceptron (MLP) feature mapping function; Design a triangle feature map:
$f_{i j k}=\psi\left(e_{i j}, e_{i k}, \theta_{i j k}\right)
$ Where $\psi$ is the triangle feature mapping function, combining two edge vectors $e_{i j}, e_{i k}$ and angle information $\theta_{i j k}$, Complete the triangular local feature modeling. Triangle local attention weight calculation.
\vspace{-3mm}
\begin{equation}
\begin{aligned}
\label{eq:TWLA}
A_{i j}=\operatorname{Softmax}\left(\frac{Q_i K_j^T}{\sqrt{D}}+G_{i j}\right),
\end{aligned}
\vspace{-2mm}
\end{equation}
Finally, the triangle local feature aggregation is performed
\vspace{-3mm}
\begin{equation}
\begin{aligned}
\label{eq:TWLA1}
Y_i=\sum_{j \in \mathcal{N}(i)} \sum_{k \in \mathcal{N}(j)} G_{i j k} A_{i j} V_j.
\end{aligned}
\vspace{-2mm}
\end{equation}
\vspace{-1mm}
The local relationship is modeled through the triangle structure, emphasizing the geometric information between points and neighbors, and the computational complexity is reduced compared to global attention.

Finally, we concatenate the rectangular window-based CGA with our proposed triangular TWLA and seamlessly combine them in a Transformer Layer to further improve the performance of image restoration MatIR. The advantage of designing the Transformer layer in this way is that the model can activate more input pixels from the local and global, triangular attention window and rectangular attention window range to achieve higher quality image restoration.

\begin{figure*}
    \centering
    
    \captionsetup{font={small}} 
    
    \includegraphics[width=.99\linewidth]{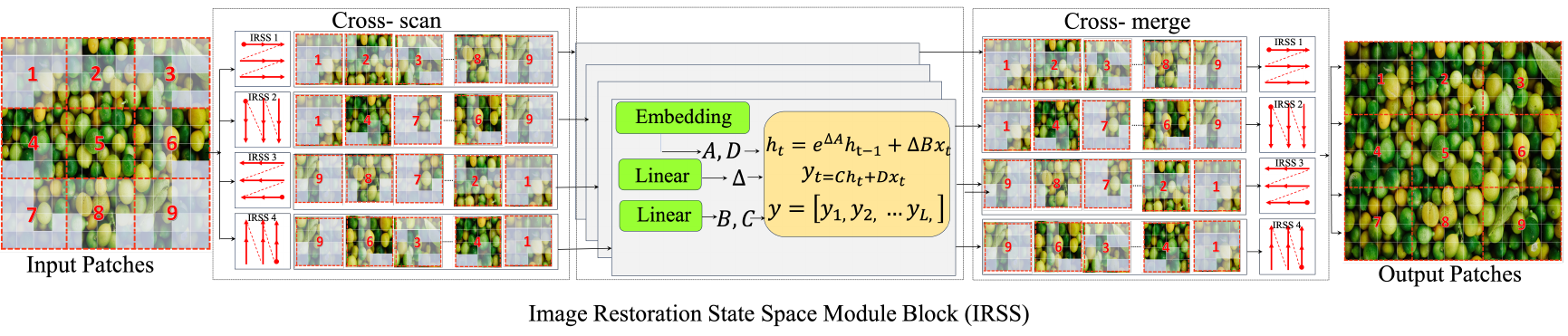}
    \vspace{-3mm}
    \caption{More details on IRSS, the core component of our MatIR model.}
    \label{fig:More details on IRSS}
    \vspace{-6mm}
\end{figure*}

\subsection{Image Restoration State Space Block}
The computational cost of the Transformer attention mechanism grows quadratically with the sequence length, which limits their scalability, especially for high-resolution image restoration tasks. The state-space model (SSM) uses state-space representation to achieve linear computational complexity, which can efficiently process long sequences without compromising accuracy. Inspired by these characteristics, we introduced the image restoration state-space module in the restoration model design (see \ref{sec:SSM_Module}and \cref{fig:More details on IRSS}), and in order to maximize the performance of this module, IRSS creates linear computational complexity information related to the sequence length in the state space from four different directional paths: left, top, right, and bottom. The designed IRSS block acts as an independent Mamba layer of the MatIR model to interact with the Transformer layer.
\vspace{-3mm}
\subsection{MatIR Overall Architecture}
\vspace{-1mm}
As shown in \cref{fig:MatIR1}, our MatIR consists of three stages: shallow feature extraction, deep feature extraction, and high-quality reconstruction. Given a low-quality (LQ) input image $I_{LQ} \in \mathds{R}^{H \times W \times 3}$, we first employ a $3\times 3$ convolution layer from the shallow feature extraction to generate the shallow feature $F_S \in \mathds{R}^{H \times W \times C}$, where $H$ and $W$ represent the height and width of the input image, and $C$ is the number of channels. Subsequently, the shallow feature $F_S$ undergoes the deep feature extraction stage to acquire the deep feature $F_D^l \in \mathds{R}^{H \times W \times C}$ at the $l$-th layer, $l \in \{1,2,\cdots L\}$. The deep learning feature extraction stage uses Transformer and Mamba stacked layers. In the Transformer layer, we use triangular window-based local attention (TWLA) and channel-based global attention (CGA) to effectively activate the attention mechanism within a wider range of image pixels to improve the performance of this module. In the Mamba layer, we use the Image Inpainting State Space (IRSS) module, which traverses along four scan paths to achieve efficient processing of long sequence data from different directions and paths. Improves performance and throughput while maintaining a manageable memory footprint. As the three core components of our MatIR, TWLA and CGA each attentionally activate more input pixels from the range of local and global, triangular and rectangular windows to achieve higher quality image recovery. IRSS creates linear computational complexity information related to the sequence length in the state space from four different direction paths: left, up, right, and down. It shows efficiency and accuracy in handling long-distance dependencies in sequence modeling tasks. Effect. Finally, we use the element-wise sum to obtain the input of the high-quality reconstruction stage $F_R = F^L_D + F_S$, which is used to reconstruct the high-quality (HQ) output image $I_{HQ}$.

%% file: sec/4_exp.tex
\begin{figure}[!tb]  
\begin{minipage}[t]{0.99\linewidth}
        \centering
        \captionsetup{width=0.99\linewidth}
        \captionof{table}{Ablation experiments for different design choices of MatIR.}
        \vspace{-5mm}
       \label{tab:ablation-irssb}
        \setlength{\tabcolsep}{1.8pt}
        \scalebox{0.8}{
       \begin{tabular}{l|cccccc}
\toprule
settings        & Set5  & Set14  & Urban100 \\ \midrule
(1)remove TWLA      &\rd 38.65 & \rd35.11  & \nd 35.07  \\
(2)remove CGA  &\fs 38.69 & \fs 35.13 &  \rd 35.06  \\
(3)remove IRSS  &\nd 38.67 &  \nd 35.12 & \fs 35.08 \\ \bottomrule
\end{tabular}%
        }
    \end{minipage}
\vspace{-5mm}
\end{figure}

\begin{table}
 \centering
 \caption{Model size and computational burden comparisons between MatIR and recent state-of-the-art methods.}
 \vspace{-2mm}
 \label{tab:model size comparisons}
    \captionsetup{font={small}}
    \footnotesize
    \setlength{\tabcolsep}{4pt}
    \scalebox{0.9}{

    \begin{tabular}{|l|cc|cc|cc|}
        \hline
        
        \multirow{2}{*}{\textbf{Model}} & \multirow{2}{*}{\textbf{Params}} & \multirow{2}{*}{\textbf{FLOPs}} & \multicolumn{2}{c|}{\textbf{Urban100}} & \multicolumn{2}{c|}{\textbf{Manga109}} \\
        & & & PSNR & SSIM & PSNR & SSIM \\
        \hline \hline
        CAT-A    & 16.6M & 360G & 27.89 & 0.8339 & 32.39 & 0.9285 \\
        HAT      & 20.8M & 412G &\rd  27.97 & \fs  0.9368 & \rd  32.48 & \rd  0.9292 \\
        ATD      & 20.3M & 417G & \nd 28.17 &\rd  0.8404 &\nd  32.63 &\nd  0.9306 \\
        MatIR    & 19.2M & 398G & \fs 28.57 &\nd  0.8434 & \fs 32.93 &\fs  0.9356 \\
        \hline
    
    \end{tabular}
    }
    \vspace{-4mm}
\end{table}

\begin{table}[t]
	\centering
    \vspace{1mm}
    \caption{Running time of different methods. $*$-T: T timesteps.}
    \vspace{-1mm}
	\label{tab:runtime}
    \vspace{-1mm}
	\scriptsize
    \resizebox{1\columnwidth}{!}{	
	\begin{tabular}{cccccccc}
		\toprule
        \!\!\!\!Methods\!\!\!\! & \!\!\!\!CAT-A \cite{chen2022cross}\!\!\!\! & \!\!\!\!HAT \cite{chen2023hat}\!\!\!\! & \!\!\!\!ATD \cite{waleffe2024empirical}\!\!\!\! & \!\!\!\!Ours-10\!\!\!\! & \!\!\!\!Ours-15\!\!\!\! & \!\!\!\!Ours-20\!\!\!\! \\	
        \midrule   
        {\textbf{Time (s)}} 
        & 15.58 & \nd15.53 & \rd 15.55 & \fs  15.32 & 17.63 & 22.18 \\
        \rule{0pt}{5pt}
        {\textbf{PSNR↑}} 
        & 38.45 & 38.39 & 38.44 & \rd 38.58 & \nd 39.25 & \fs 39.47 \\
		\bottomrule
	\end{tabular}
    }
    \vspace{-3mm}
	
\end{table}

\begin{table*}[!t]
\centering
\caption{Quantitative comparison on \underline{\textbf{classic image super-resolution}} with state-of-the-art methods. Best results are highlighted as \colorbox{colorFst}{\fs \!first\!}, \colorbox{colorSnd}{\!second\!} and \colorbox{colorTrd}{\!third\!}.}
\label{tab:classicSR}
\setlength{\tabcolsep}{2pt}
\scalebox{0.8}{
\begin{tabular}{@{}l|c|cc|cc|cc|cc|cc@{}}
\toprule
 & & \multicolumn{2}{c|}{\textbf{Set5}} &
  \multicolumn{2}{c|}{\textbf{Set14}} &
  \multicolumn{2}{c|}{\textbf{BSDS100}} &
  \multicolumn{2}{c|}{\textbf{Urban100}} &
  \multicolumn{2}{c}{\textbf{Manga109}} \\
\multirow{-2}{*}{Method} & \multirow{-2}{*}{scale} & PSNR  & SSIM   & PSNR  & SSIM   & PSNR  & SSIM   & PSNR  & SSIM   & PSNR  & SSIM   \\ \midrule
EDSR~\cite{lim2017enhanced}   & $\times 2$ & 38.11 & 0.9602 & 33.92 & 0.9195 & 32.32 & 0.9013 & 32.93 & 0.9351 & 39.10 & 0.9773 \\
RCAN~\cite{zhang2018image}   & $\times 2$ & 38.27 & 0.9614 & 34.12 & 0.9216 & 32.41 & 0.9027 & 33.34 & 0.9384 & 39.44 & 0.9786 \\
SAN~\cite{dai2019second}    & $\times 2$ & 38.31 & 0.9620 & 34.07 & 0.9213 & 32.42 & 0.9028 & 33.10 & 0.9370 & 39.32 & 0.9792 \\
HAN~\cite{niu2020single}    & $\times 2$ & 38.27 & 0.9614 & 34.16 & 0.9217 & 32.41 & 0.9027 & 33.35 & 0.9385 & 39.46 & 0.9785 \\
IGNN~\cite{zhou2020cross} & $\times$2 &38.24 & 0.9613 & 34.07 & 0.9217 & 32.41 & 0.9025 & 33.23 & 0.9383 & 39.35 & 0.9786\\
CSNLN~\cite{mei2020image}  & $\times 2$ & 38.28 & 0.9616 & 34.12 & 0.9223 & 32.40 & 0.9024 & 33.25 & 0.9386 & 39.37 & 0.9785 \\
NLSA~\cite{mei2021image}   & $\times 2$ & 38.34 & 0.9618 & 34.08 & 0.9231 & 32.43 & 0.9027 & 33.42 & 0.9394 & 39.59 & 0.9789 \\
ELAN~\cite{zhang2022efficient}   & $\times 2$ & 38.36 & 0.9620 & 34.20 & 0.9228 & 32.45 & 0.9030 & 33.44 & 0.9391 & 39.62 & 0.9793 \\
IPT~\cite{chen2021pre} & $\times 2$ & 38.37 & - &34.43 &-& 32.48&-& 33.76& -& -& - \\
SwinIR~\cite{liang2021swinir} & $\times 2$ &38.42 & 0.9623 & 34.46 &  0.9250 & 32.53 &  0.9041 &  33.81 & 0.9427 &  39.92 &  0.9797 \\
SRFormer~\cite{zhou2023srformer} & $\times 2$ &  38.51 & 0.9627& 34.44& 0.9253& 32.57 &0.9046 & 34.09&  {0.9449} & 40.07 &  0.9802\\
CAT-A~\cite{chen2022cross} & $\times 2$ &     38.51 &   0.9626&    34.78&   0.9265&   32.59 &   0.9047 &   34.26&    {0.9440} &   40.10 &   0.9805\\
MambaIR~\cite{guo2025mambair} & $\times 2$ &   38.57 & 0.9627& 34.67& 0.9261& 32.58 & 0.9048 & 34.15&    {0.9446} &   40.28 &   0.9806\\
HAT~\cite{chen2023hat} & $\times 2$ &  \rd 38.63 &  \rd 0.9630&  34.86&   0.9274&   32.62 &   0.9053 &   34.45&   {0.9466} &   40.26 &   0.9809\\
ATD~\cite{zhang2024transcending}   & $\times 2$ & 38.61 & 0.9629 & \rd 34.95 & \rd 0.9276 & \rd 32.65 & \rd 0.9056 & \rd 34.70 & \rd 0.9476 & \rd 40.37 & \rd 0.9810 \\
GRL~\cite{li2023efficient}   & $\times 2$ & \nd 38.67 & \nd 0.9647 & \nd 35.08 & \nd 0.9303 & \nd 32.67 & \nd 0.9087 & \nd 35.06 & \nd 0.9505 & \nd 40.67 & \nd 0.9818 \\
{\textbf{MatIR (Ours)}}   & $\times 2$ &\fs{38.70}&\fs{0.9648}&\fs{35.13}&\fs {0.9304}&\fs{32.73}&\fs{0.9048}&\fs{35.11}& \fs {0.9507}&\fs{40.33}&\fs{0.9806} \\     \midrule
EDSR~\cite{lim2017enhanced} & $\times$3 & 
34.65 & 0.9280 & 30.52 & 0.8462 & 29.25 & 0.8093 & 28.80 & 0.8653 & 34.17 & 0.9476\\
RCAN~\cite{zhang2018image} & $\times$3 &
34.74 & 0.9299 & 30.65 & 0.8482 & 29.32 & 0.8111 & 29.09 & 0.8702 & 34.44 & 0.9499\\
SAN~\cite{dai2019second} & $\times$3 &
34.75 & 0.9300 & 30.59 & 0.8476 & 29.33 & 0.8112 & 28.93 & 0.8671 & 34.30 & 0.9494\\
HAN~\cite{niu2020single} & $\times$3 &
34.75 & 0.9299 & 30.67 & 0.8483 & 29.32 & 0.8110 & 29.10 & 0.8705 & 34.48 & 0.9500\\
IGNN~\cite{zhou2020cross} & $\times$3 &
34.72 & 0.9298 & 30.66 & 0.8484 & 29.31 & 0.8105 & 29.03 & 0.8696 & 34.39 & 0.9496\\
CSNLN~\cite{mei2020image} & $\times$3 &
34.74 & 0.9300 & 30.66 & 0.8482 & 29.33 & 0.8105 & 29.13 & 0.8712 & 34.45 & 0.9502\\
NLSA~\cite{mei2021image} & $\times$3 &
34.85 & 0.9306 & 30.70 & 0.8485 & 29.34 & 0.8117 & 29.25 & 0.8726 & 34.57 & 0.9508\\
ELAN~\cite{zhang2022efficient} & $\times$3 &
34.90 & 0.9313 & 30.80 & 0.8504 & 29.38 & 0.8124 & 29.32 & 0.8745 & 34.73 & 0.9517\\
IPT~\cite{chen2021pre} & $\times$3 &
34.81 & - & 30.85 & - & 29.38 & - & 29.49 & - & - & - \\
SwinIR~\cite{liang2021swinir} & $\times$3 &
34.97 &  0.9318 &  30.93 &  0.8534 &  29.46 &  0.8145 &  29.75 &  0.8826 &  35.12 &  0.9537\\
SRformer~\cite{zhou2023srformer} & $\times$3 &
35.02&  0.9323 &  30.94 &  {0.8540} &  29.48 &  0.8156 & {30.04} &  0.8865 &  35.26 &  0.9543\\
CAT-A~\cite{chen2022cross} & $\times$3 &
   35.06&    0.9326 &    31.04 &    {0.8538} &    29.52 &    0.8160 &   {30.12} &    0.8862 &    35.38 &    0.9546\\
MambaIR~\cite{guo2025mambair} & $\times$3 &
   \rd 35.08&    0.9323 &    30.99 &    {0.8536} &    29.51 &    0.8157 &   {29.93} &    0.8841 &    35.43 &    0.9546\\
HAT~\cite{chen2023hat} & $\times$3 &
   35.07&    \nd 0.9329 &    \rd 31.08 &    \nd{0.8555} &    \rd 29.54 &    \nd 0.8167 &   \nd{30.23} &    \nd 0.8896 &    \nd 35.53 &    \nd 0.9552\\
ATD~\cite{zhang2024transcending} & $\times$3 &
\nd 35.11 & \fs 0.9330 & \fs 31.13 & \fs 0.8556 & \fs 29.57 & \fs 0.8176 & \fs 30.46 & \fs 0.8917 & \fs 35.63 & \fs 0.9558\\
{\textbf{MatIR (Ours)}}  & $\times$3 &  \fs{35.13} & \rd{0.9328} &  \nd{31.06} & \nd{0.8555} & \nd{29.56} & \rd{0.8163} & \nd{30.23} & \rd{0.8888} & \rd{35.47} & \rd{0.9551}
\\ \midrule
EDSR~\cite{lim2017enhanced}   & $\times 4$ & 32.46 & 0.8968 & 28.80 & 0.7876 & 27.71 & 0.7420 & 26.64 & 0.8033 & 31.02 & 0.9148 \\
RCAN~\cite{zhang2018image}     & $\times 4$ & 32.63 & 0.9002 & 28.87 & 0.7889 & 27.77 & 0.7436 & 26.82 & 0.8087 & 31.22 & 0.9173 \\
SAN~\cite{dai2019second}      & $\times 4$ & 32.64 & 0.9003 & 28.92 & 0.7888 & 27.78 & 0.7436 & 26.79 & 0.8068 & 31.18 & 0.9169 \\
HAN~\cite{niu2020single}      & $\times 4$ & 32.64 & 0.9002 & 28.90 & 0.7890 & 27.80 & 0.7442 & 26.85 & 0.8094 & 31.42 & 0.9177 \\
IGNN~\cite{zhou2020cross}     & $\times$4 &32.57 & 0.8998 & 28.85 & 0.7891 & 27.77 & 0.7434 & 26.84 & 0.8090 & 31.28 & 0.9182\\
CSNLN~\cite{mei2020image}     & $\times 4$ & 32.68 & 0.9004 & 28.95 & 0.7888 & 27.80 & 0.7439 & 27.22 & 0.8168 & 31.43 & 0.9201 \\
NLSA~\cite{mei2021image}      & $\times 4$ & 32.59 & 0.9000 & 28.87 & 0.7891 & 27.78 & 0.7444 & 26.96 & 0.8109 & 31.27 & 0.9184 \\
ELAN~\cite{zhang2022efficient}   & $\times 4$ & 32.75 & 0.9022 & 28.96 & 0.7914 & 27.83 & 0.7459 & 27.13 & 0.8167 & 31.68 & 0.9226 \\
IPT ~\cite{chen2021pre}          & $\times 4$ & 32.64 &-& 29.01 &- &27.82 &-& 27.26& -& -& - \\
SwinIR~\cite{liang2021swinir}    & $\times 4$ &  32.92 &  0.9044 &  29.09 &  0.7950 &  27.92 &  0.7489 &  27.45 &  0.8254 &  32.03 &  0.9260 \\
SRFormer~\cite{zhou2023srformer} & $\times 4$ &  32.93 &  0.9041 &  29.08 &  0.7953 & 27.94 &  0.7502 &  27.68 & {0.8311} &   32.21 &   0.9271 \\
CAT-A~\cite{chen2022cross} & $\times 4$ &    33.08 &     0.9052 &     29.18 &    0.7960 &   27.99 &    0.7510 &    27.89 &   {0.8339} &     32.39 &     0.9285 \\
MambaIR~\cite{guo2025mambair} & $\times 4$ &    33.03 &     0.9046 &     29.20 &    0.7961 &   27.98 &    0.7503 &    27.68 &   {0.8287} &     32.32 &     0.9272 \\
HAT~\cite{chen2023hat} & $\times 4$ &    33.04 &     0.9056 &     29.23 &    0.7973 &   28.00 &    0.7517 &    27.97 &   {0.8368} &     32.48 &     0.9292 \\
ATD~\cite{zhang2024transcending}   & $\times 4$ & \nd 33.10 & \nd 0.9058 & \rd 29.24 & \rd 0.7974 & \nd 28.01 &\rd 0.7526 & \rd 28.17 & \rd 0.8404 & \rd 32.62 & \rd 0.9306 \\
GRL~\cite{li2023efficient}   & $\times 4$ & \nd 33.10 & \fs 0.9094 & \nd 29.37 & \nd 0.8058 & \nd 28.01 & \fs 0.7611 & \nd 28.53 & \nd 0.8504 & \nd 32.77 & \nd 0.9325 \\
{\textbf{MatIR (Ours)}}   & $\times 4$ &\fs{33.14}&\rd {0.9055}&\fs{29.40}&\fs{0.8059}&\fs{28.03}&\nd{0.7610}&\fs{28.55}&\fs {0.8505}&\fs{32.82}&\fs{0.9326} \\  \bottomrule
\end{tabular}%
}
  \vspace{-4mm}
\end{table*}

    
    

\begin{figure}[t]
  \hspace{-10pt}
   \includegraphics[width=1.08\linewidth]{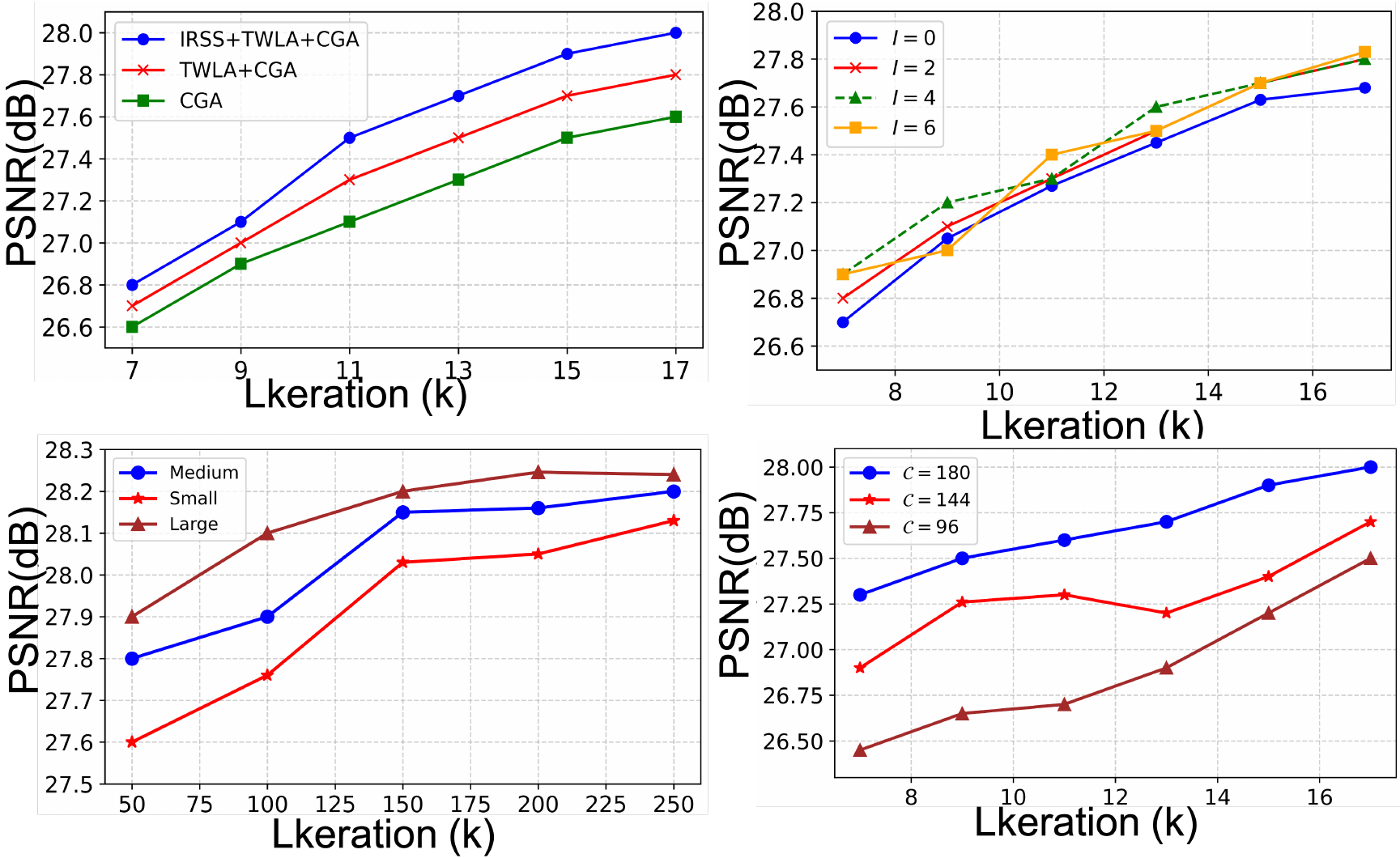}
   \vspace{-8mm}
   \caption{Comparison of iterative performance (PSNR in dB) of the proposed MatIR \textbf{Top-Left:} Performance comparison of triangle local attention, channel global attention and image recovery space modules. \textbf{Top-Right:} Triangle local attention and channel global attention, various interval sizes, \textbf{Bottom left:}
Different channel lengths. [On BSD100($\times4$), epoch 70]and \textbf{bottom right:}Performance evaluation of small, medium and large MatIR models.}
   \vspace{-4mm}
   \label{fig:image_2}
\end{figure}

\begin{figure}[t]
  \hspace{-10pt}
   \includegraphics[width=1.08\linewidth]{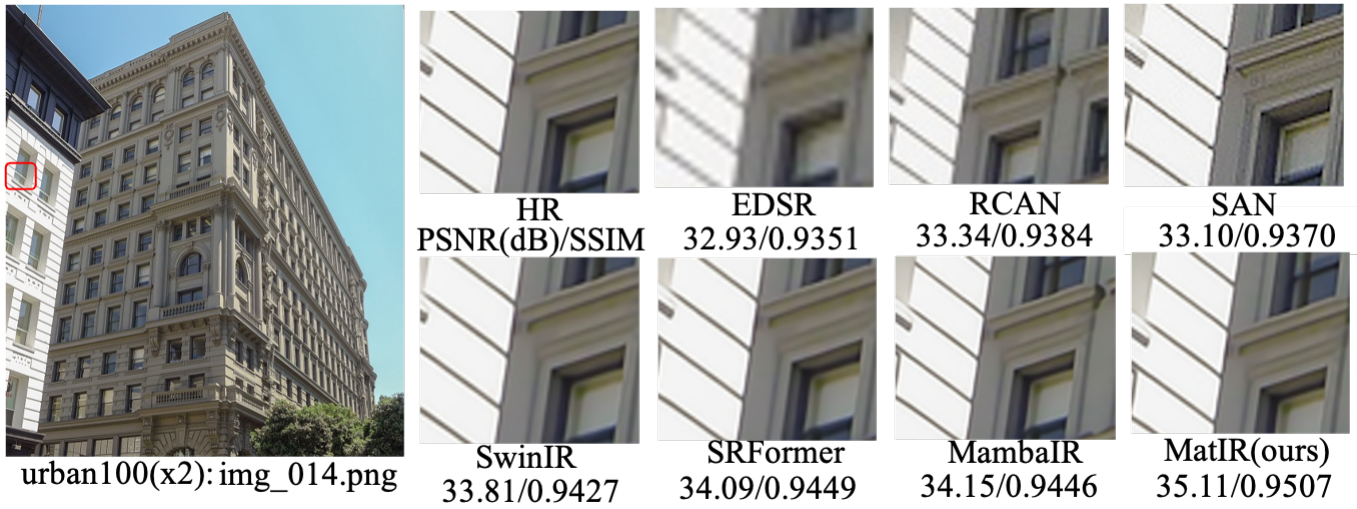}
   \vspace{-7mm}
   \caption{The visual comparison of the MatIR network on x2SR utilizes red bounding boxes to highlight the patch for comparison, in order to better reflect performance differences.}
   \vspace{-4mm}
   \label{fig:srx2}
\end{figure}



\vspace{-2mm}
\begin{table}[t]
\centering
\caption{ \underline{\textbf{gaussian color image denoising}} Quantitative comparison with state-of-the-art methods.}
\label{tab:guassian-denoise}
\setlength{\tabcolsep}{2pt}
\scalebox{0.65}{
\begin{tabular}{l|ccc|ccc|ccc|ccc}
\toprule
\multirow{2}{*}{Method} &  \multicolumn{3}{c|}{\textbf{BSD68}} & \multicolumn{3}{c|}{\textbf{Kodak24}} & \multicolumn{3}{c|}{\textbf{McMaster}} & \multicolumn{3}{c}{\textbf{Urban100}}\\
&$\sigma$=15 & $\sigma$=25 & $\sigma$=50 &$\sigma$=15 & $\sigma$=25 & $\sigma$=50 &$\sigma$=15 & $\sigma$=25 & $\sigma$=50 &$\sigma$=15 & $\sigma$=25 & $\sigma$=50\\
\midrule
IRCNN~\cite{IRCNN} &
33.86 & 31.16 & 27.86 & 34.69 & 32.18 & 28.93 & 34.58 & 32.18 & 28.91 & 33.78 & 31.20 & 27.70\\
FFDNet~\cite{FFDNet} &
33.87 & 31.21 & 27.96 & 34.63 & 32.13 & 28.98 & 34.66 & 32.35 & 29.18 & 33.83 & 31.40 & 28.05\\
DnCNN~\cite{DnCNN} &
33.90 & 31.24 & 27.95 & 34.60 & 32.14 & 28.95 & 33.45 & 31.52 & 28.62 & 32.98 & 30.81 & 27.59\\
DRUNet~\cite{DRUNet}&
34.30 & 31.69 & 28.51 & 35.31 & 32.89 & 29.86 & 35.40 & 33.14 & 30.08 & 34.81 & 32.60 & 29.61\\
SwinIR~\cite{liang2021swinir} &
\rd{34.42} &  31.78 & 28.56 &  35.34 & 32.89 &  29.79 &\rd 35.61 &  33.20 &  30.22 & \rd 35.13 &  32.90 &  29.82\\
Restormer~\cite{zamir2022restormer} &  34.40 & \rd{31.79} & \rd{28.60} & \nd {35.47} & \fs {33.04} & \fs {30.01} & \rd{35.61} & \rd{33.34} & \rd{30.30} & \rd{35.13} & \rd{32.96} & \rd{30.02} \\ 
MambaIR~\cite{guo2025mambair} & \nd 34.48 & \nd{32.24} & \nd{28.66} & \rd {35.42} & \nd {32.99} & \rd {29.92} & \nd{35.70} & \nd{33.43} & \nd{30.35} & \nd{35.37} & \nd{33.21} & \nd{30.30} \\ 
{\textbf{MatIR (Ours)}}  &
\fs {34.51} & \fs {32.28} & \fs {28.69} & \fs{35.51} & \nd{32.99} & \nd{29.98} & \fs {35.73} &\fs {33.45} & \fs {30.40} & \fs {35.39} & \fs {33.25} & \fs {30.33}\\
\bottomrule
\end{tabular} 
}
\vspace{-3mm}
\end{table}

\vspace{-11mm}
\begin{figure}[t]
  \hspace{-10pt}
   \includegraphics[width=1.08\linewidth]{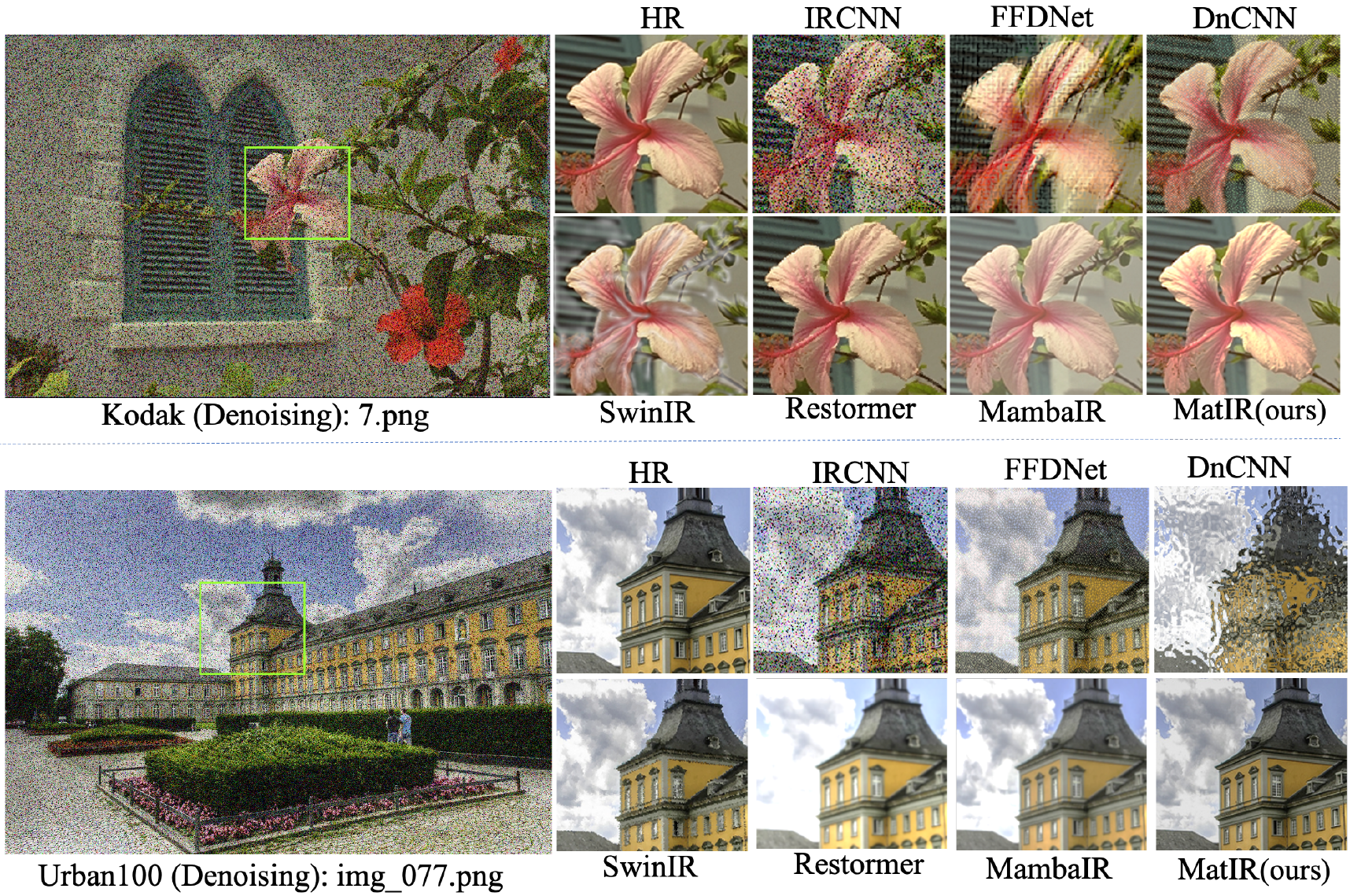}
   \vspace{-7mm}
   \caption{Visual comparisons between MatIR and state-of-the-art Denoising methods.}
   \vspace{-6mm}
   \label{fig:x2sr}
\end{figure}


\vspace{12mm}
\section{Experiments}
\vspace{-1mm}
\subsection{Experiment settings.}

\noindent \textbf{Dataset and Evaluation.} 
Following the setup in previous works~\cite{liang2021swinir,zhang2023accurate}, we conduct experiments on various image restoration tasks, including image super-resolution and image denoising (\textit{i.e.}, Gaussian color image denoising and real-world denoising), and Defocus deblurring results. (\textit{i.e.}, S: single-image defocus deblurring. D: dual-pixel defocus deblurring). We employ DIV2K~\cite{timofte2017ntire} and Flickr2K~\cite{lim2017enhanced} to train classic SR models . Moreover, we use Set5~\cite{bevilacqua2012low}, Set14~\cite{zeyde2012single}, B100~\cite{martin2001database}, Urban100~\cite{huang2015single}, and Manga109~\cite{matsui2017sketch} to evaluate the effectiveness of different SR methods. For gaussian color image denoising, we utilize DIV2K~\cite{timofte2017ntire}, Flickr2K~\cite{lim2017enhanced}, BSD500~\cite{arbelaez2010contour}, and WED~\cite{ma2016waterloo} as our training datasets. Our testing datasets for guassian color image denoising includes BSD68~\cite{martin2001database}, Kodak24~\cite{kao}, McMaster~\cite{zhang2011color}, and Urban100~\cite{huang2015single}. For real image denoising, 
we train our model with 320 high-resolution images from SIDD~\cite{abdelhamed2018high} datasets, and use the SIDD test set and DND ~\cite{plotz2017benchmarking} dataset for testing. 
Following~\cite{liang2021swinir,zhang2018residual}, we denote the model as MatIR when self-ensemble strategy~\cite{lim2017enhanced} is used in testing. The performance is evaluated using PSNR and SSIM on the Y channel from the YCbCr color space. Defocus Deblurring: We compare against 12 state-of-the-art methods on datasets comprising outdoor, indoor, and mixed scenes ~\cite{karaali2017edge,  son2021single,abuolaim2020defocus,wang2022uformer,lee2021iterative,zamir2022restormer,li2023efficient}.

\vspace{1mm}
\noindent \textbf{Training Details.} 
In accordance with previous works~\cite{liang2021swinir,chen2023activating,zhang2023accurate}, we perform data augmentation by applying horizontal flips and random rotations of $90^\circ, 180^\circ$, and $270^\circ$. Additionally, we crop the original images into $64 \times 64$ patches for image SR and $128 \times 128$ patches for image denoising during training. For image SR, we use the pre-trained weights from the $\times$2 model to initialize those of $\times$3 and $\times$4 and halve the learning rate and total training iterations to reduce training time~\cite{lim2017enhanced}. To ensure a fair comparison, we adjust the training batch size to 32 for image SR and 16 for image denoising. We employ the Adam~\cite{kingma2014adam} as the optimizer for training our MatIR with $\beta_1 = 0.9, \beta_2 = 0.999$. The initial learning rate is set at $2 \times 10^{-4}$ and is halved when the training iteration reaches specific milestones. Our MatIR model is trained with 8 NVIDIA A100 GPUs.

\noindent\textbf{Evaluation metrics.}
We use PSNR, SSIM, and LPIPS as the evaluation metrics for most image restoration tasks. In general, higher PSNR and SSIM, and lower LPIPS and FID mean better performance.

\vspace{0.1mm}
\subsection{Ablation Study}
\vspace{-1mm}
Impact of different designs of Transformer Layer. As core components, TWLA and CGA can achieve higher quality image restoration for MatIR by activating more input pixels for local and global, triangular attention window and rectangular attention window ranges. In this section, we perform ablation studies on these two key components respectively. The results shown \cref{tab:ablation-irssb}, \cref{tab:model size comparisons}: (1) Concatenating the two components in the Transformer Layer has a greater gain effect than using each component separately. (2) If the core component IRSS of the Mamba layer uses the Transformer method, the computational requirements will increase with the same benefit. Impact of different scanning modes in IRSS. In order for Mamba to process 2D images, the feature map needs to be flattened first and then iterated by the state space equation. Therefore, the unfolding strategy is particularly important. In this work, we follow ~\cite{liu2024vmamba} to generate scanning sequences using four different scanning directions. Here we ablate different scanning modes to study the effect, Compared with single-direction (upper left to lower right) and bidirectional (upper left to lower right, lower right to upper left) scanning, using four-directional scanning allows the anchor pixel to perceive a larger range of neighborhoods, thus achieving better results.

\begin{table*}[!t]
\begin{center}
\vspace{-3mm}

\caption{\textit{\textbf\\{Defocus deblurring}} results. \textbf{S:} single-image defocus deblurring. \textbf{D:} dual-pixel defocus deblurring.}
\label{table:defocus_deblurring}  
\setlength{\tabcolsep}{1.9pt}
\scalebox{0.75}{
\begin{tabular}{l | c | c | c | c | c | c | c | c | c | c | c | c}
\toprule[0.1em]
\multirow{2}{*}{Method} & \multicolumn{4}{c|}{\textbf{Indoor Scenes}} & \multicolumn{4}{c|}{\textbf{Outdoor Scenes}} & \multicolumn{4}{c}{\textbf{Combined}} \\ \cline{2-13}
	&PSNR$\uparrow$ & SSIM$\uparrow$ & MAE$\downarrow$ & LPIPS$\downarrow$			&PSNR$\uparrow$ & SSIM$\uparrow$ & MAE$\downarrow$ & LPIPS$\downarrow$ &PSNR$\uparrow$ & SSIM$\uparrow$ & MAE$\downarrow$ & LPIPS$\downarrow$		\\ \midrule[0.1em]
EBDB$_S$~\cite{karaali2017edge}	&25.77	&0.772	&0.040	&0.297	&21.25	&0.599	&0.058	&0.373	&23.45	&0.683	&0.049	&0.336	\\
DMENet$_S$~\cite{lee2019deep}	&25.50	&0.788	&0.038	&0.298	&21.43	&0.644	&0.063	&0.397	&23.41	&0.714	&0.051	&0.349	\\
JNB$_S$~\cite{shi2015just}	&26.73	&0.828	&0.031	&0.273	&21.10	&0.608	&0.064	&0.355	&23.84	&0.715	&0.048	&0.315	\\
DPDNet$_S$~\cite{abuolaim2020defocus}	&26.54	&0.816	&0.031	&0.239	&22.25	&0.682	&0.056	&0.313	&24.34	&0.747	&0.044	&0.277	\\
KPAC$_S$~\cite{son2021single}	&27.97	&0.852	&0.026	&0.182	&22.62	&0.701	&0.053	&0.269	&25.22	&0.774	&0.040	&0.227	\\
IFAN$_S$~\cite{lee2021iterative}	&\rd 28.11	&\rd 0.861	&\rd 0.026	&\rd 0.179	&\rd 22.76	&\rd 0.720	&\rd 0.052	&\rd 0.254	&\rd 25.37	&\rd 0.789	&\rd 0.039	&\rd 0.217	\\
Restormer$_S$~\cite{zamir2022restormer}	&\nd {28.87}	&\nd {0.882}	&\nd {0.025}	&\nd {0.145}	&\nd {23.24}	&\nd {0.743}	&\nd {0.050}	&\nd {0.209}	&\nd {25.98}	&\nd {0.811}	&\nd {0.038}	&\nd {0.178}	\\
{\textbf{MatIR$_S$-B (Ours)}}	&\fs {29.23}	&\fs{0.891}	&\fs{0.021}	&\fs{0.129}	&\fs{23.62}	&\fs{0.783}	&\fs{0.045}	&\fs{0.159}	&\fs{26.79}	&\fs{0.841}	&\fs{0.031}	&\fs{0.143}	
\\ \midrule[0.1em]
DPDNet$_D$~\cite{abuolaim2020defocus}	&27.48	&0.849	&0.029	&0.189	&22.90	&0.726	&0.052	&0.255	&25.13	&0.786	&0.041	&0.223	\\
RDPD$_D$~\cite{abuolaim2021learning}	&28.10	&0.843	&0.027	&0.210	&22.82	&0.704	&0.053	&0.298	&25.39	&0.772	&0.040	&0.255	\\
Uformer$_D$~\cite{wang2022uformer}	&28.23	&0.860	&0.026	&0.199	&23.10	&0.728	&0.051	&0.285	&25.65	&0.795	&0.039	&0.243	\\
IFAN$_D$~\cite{lee2021iterative}	&\rd 28.66	&\rd 0.868	&\rd 0.025	&\rd 0.172	&\rd 23.46	&\rd 0.743	&\rd 0.049	&\rd 0.240	&\rd 25.99	&\rd 0.804	&\rd 0.037	&\rd 0.207	\\
Restormer$_D$~\cite{zamir2022restormer}	&\nd {29.48}	&\nd {0.895}	&\nd {0.023}	&\nd {0.134}	&\nd {23.97}	&\nd {0.773}	&\nd {0.047}	&\nd {0.175}	&\nd {26.66}	&\nd {0.833}	&\nd {0.035}	&\nd {0.155}	\\
 {\textbf{MatIR$_D$-B (Ours)}} 	&\fs{30.16}	&\fs{0.911}	&\fs{0.019}	&\fs{0.093}	&\fs{24.70}	&\fs{0.817}	&\fs{0.039}	&\fs{0.123}	&\fs{27.83}	&\fs{0.862}	&\fs{0.029}	&\fs{0.105}	\\
\bottomrule[0.1em]
\end{tabular}}
\end{center}
\vspace{-10mm}
\end{table*}

\vspace{-1mm}
\subsection{Evaluation on Image Super-Resolution}
\vspace{-1mm}
we compare our method with 16 state-of-the-art IR methods on 5 public datasets of classic super-resolution.
The quantitative results are shown in \cref{tab:classicSR}.
We can see that our method outperforms most of the methods on 5 different datasets.
In particular, compared with the method SRformer, our method on Urban100 X4 leads SRformer by up to \textbf{0.87} on PSNR and leads DDNM by up to \textbf{0.0194dB} on SSIM.
For qualitative results, our method has the best visual quality, including more realistic textures, as shown in \cref{fig:srx2}.
These visual comparisons are consistent with the quantitative results, demonstrating the effectiveness of our method.
More visual results are placed in the supplementary material.

\vspace{-1mm}
\subsection{Evaluation on Image Denoising}
\vspace{-1mm}
The Gaussian color image denoising results are shown in  \cref{tab:guassian-denoise}. Similar to ~\cite{DnCNN,DRUNet}, the compared noise levels include 15, 25 and 50. It can be seen that our model achieves the best performance on most datasets. In particular, it surpasses SwinIR ~\cite{liang2021swinir} and even reaches 0.51dB
$\sigma$=50 on the Urban100 dataset. We also give a visual comparison in Figure 5. Benefiting from the global receptive field, our MatIR can achieve better structure preservation, resulting in clearer edges and natural shapes.

\begin{figure}[t]
  \hspace{-10pt}
   \includegraphics[width=1.08\linewidth]{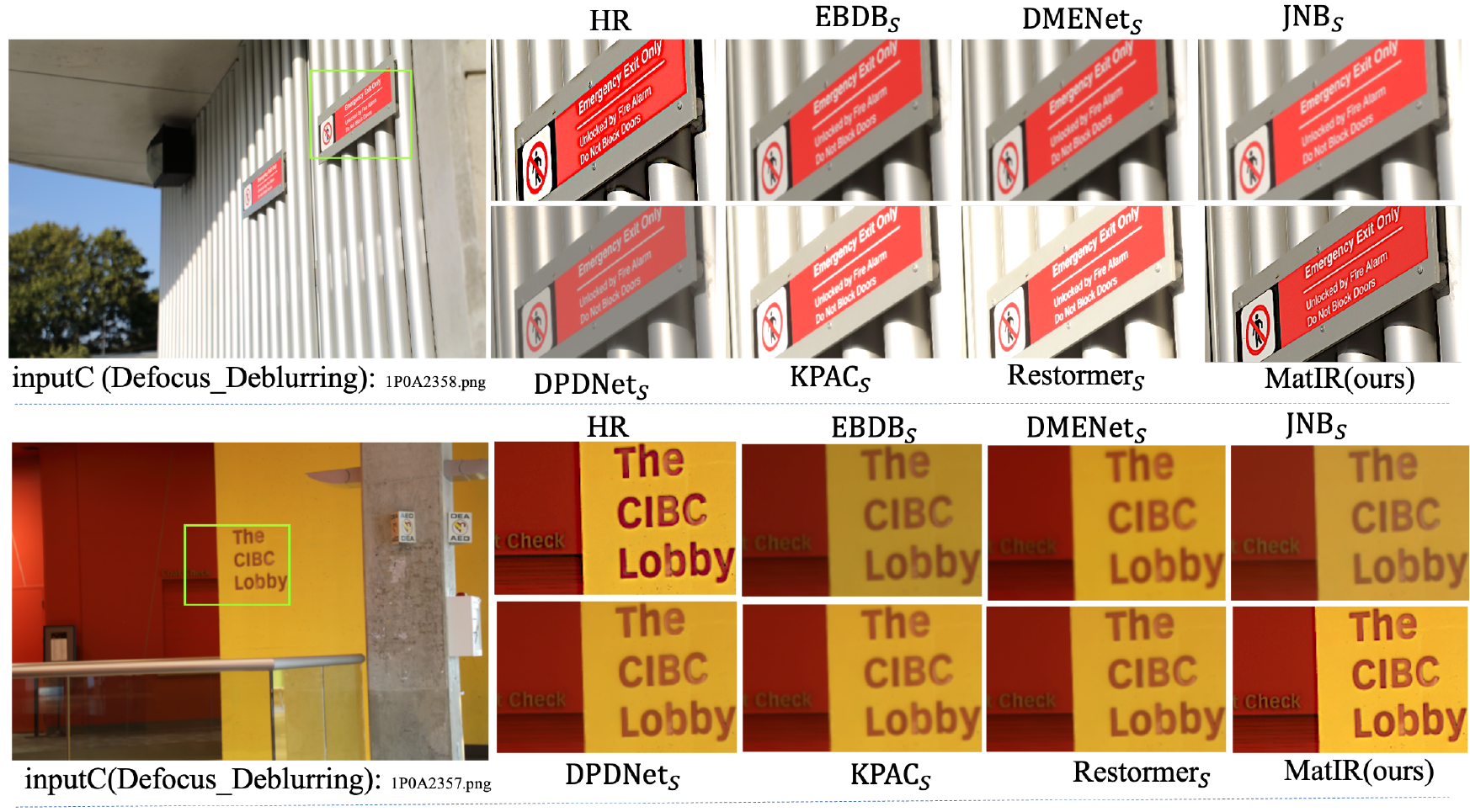}
   \vspace{-8mm}
   \caption{Visual comparisons between MatIR and state-of-the-art Defocus Deblurring methods.}
   \vspace{-6mm}
   \label{fig:x2sr}
\end{figure}


\vspace{-1mm}
\subsection{Evaluation on Image Deblurring}
\vspace{-1mm}
For image deblurring, we mainly evaluated the defocus deblurring results, including single image defocus deblurring and dual pixel defocus deblurring. In Table5, the quantitative results show that our method achieves the best performance on all datasets.
Compared with Outdoor Seenes by Restormer \cite{zamir2022restormer}, the PSRN improvement of our method can be as high as \textbf{0.73dB}.
The images we have the best visual quality, with more realistic details and close to GT images.
Due to space limitations, we provide more quantitative and qualitative results in the supplementary materials.

\vspace{-2mm}
\subsection{Real Image Denoising.}
\vspace{-2mm}
We further turn to the real image denoising task to evaluate the robustness of MatIR in the face of real-world degradation. Following~\cite{zamir2022restormer}, we adopt a progressive training strategy for fair comparison. The results are shown in \cref{tab:real_denoise}, showing that our method achieves the best performance among existing state-of-the-art models and outperforms other methods (such as Uformer~\cite{wang2022uformer}) by 0.31 dB PSNR on the SIDD dataset, demonstrating the capability of our method in real image denoising.

\vspace{1mm}
\subsection{Further Experiments}
\vspace{-2mm}
\noindent\textbf{Running time.}
We compare the running time of our proposed method with the SOTA IR method.
For a fair comparison, we evaluate all methods on $256{\times}256$ input images on NVIDIA TITAN RTX using their publicly available codes.
As shown in \cref{tab:runtime}, the running time of the proposed method is significantly better than other transformer-based methods, demonstrating the good computational efficiency of our model.

%% file: sec/5_conclusion.tex
\begin{table}[t]
\centering
\vspace{2mm}

\caption{Quantitative comparison on the \underline{\textbf{real image denosing}} task.}
\setlength{\tabcolsep}{2pt}
\scalebox{0.6}{
\begin{tabular}{l|cc|cc|cc|cc|cc|cc}
\toprule
&  \multicolumn{2}{c|}{ DeamNet~\cite{ren2021adaptive}} & \multicolumn{2}{c|}{ MPRNet~\cite{zamir2021multi} } &  \multicolumn{2}{c|}{ Uformer~\cite{wang2022uformer} }& \multicolumn{2}{c|}{ Restormer~\cite{zamir2022restormer} }& \multicolumn{2}{c|}{ MambaIR~\cite{guo2025mambair} } & \multicolumn{2}{c}{\textbf{MatIR (Ours)}}  \\
\multirow{-2}{*}{Dataset} & PSNR & SSIM & PSNR & SSIM & PSNR & SSIM & PSNR & SSIM & PSNR & SSIM & PSNR & SSIM \\
\midrule
SIDD & 39.47 & 0.957 & 39.71 & 0.958 &  {39.77} &  {0.959}& \nd {40.02} & \nd {0.960} & \rd {39.89} & \nd{0.960}& \fs{40.08} & \fs{0.963}\\
DND &  39.63 & 0.953 & 39.80 & 0.954 &  {39.96} &  {0.956} & \rd{40.03} & \nd {0.956} & \nd{40.04} & \nd{0.956}& \fs{40.05} & \fs{0.961} \\ \bottomrule
\end{tabular}%
}
 \label{tab:real_denoise}
\vspace{-5mm}
\end{table}

\vspace{-1mm}
\section{Conclusion}
\vspace{-1.5mm}
In this paper, we introduce MatIR, a novel hybrid architecture for image restoration that combines the computational efficiency of the Mamba model with the powerful contextual learning capabilities of the Transformer. By leveraging the strengths of both architectures, MatIR is able to restore high-quality images at a lower computational cost. Extensive experiments on multiple image restoration tasks demonstrate that MatIR not only achieves state-of-the-art results in terms of PSNR and SSIM, but also achieves significant improvements in terms of memory efficiency and computational complexity.
Our results suggest that hybrid architectures like MatIR offer a promising direction for advancing image restoration, especially in high-resolution and complex degraded scenes. Future work will explore extending this approach to video restoration tasks, where temporal dependencies will further benefit from the integration of SSM and Transformers.


%% file: main.bbl
\begin{thebibliography}{72}
\providecommand{\natexlab}[1]{#1}
\providecommand{\url}[1]{\texttt{#1}}
\expandafter\ifx\csname urlstyle\endcsname\relax
  \providecommand{\doi}[1]{doi: #1}\else
  \providecommand{\doi}{doi: \begingroup \urlstyle{rm}\Url}\fi

\bibitem[Abdelhamed et~al.(2018)Abdelhamed, Lin, and Brown]{abdelhamed2018high}
Abdelrahman Abdelhamed, Stephen Lin, and Michael~S Brown.
\newblock A high-quality denoising dataset for smartphone cameras.
\newblock In \emph{Proceedings of the IEEE conference on computer vision and pattern recognition}, pages 1692--1700, 2018.

\bibitem[Abuolaim and Brown(2020)]{abuolaim2020defocus}
Abdullah Abuolaim and Michael~S Brown.
\newblock Defocus deblurring using dual-pixel data.
\newblock In \emph{Computer Vision--ECCV 2020: 16th European Conference, Glasgow, UK, August 23--28, 2020, Proceedings, Part X 16}, pages 111--126. Springer, 2020.

\bibitem[Abuolaim et~al.(2021)Abuolaim, Delbracio, Kelly, Brown, and Milanfar]{abuolaim2021learning}
Abdullah Abuolaim, Mauricio Delbracio, Damien Kelly, Michael~S Brown, and Peyman Milanfar.
\newblock Learning to reduce defocus blur by realistically modeling dual-pixel data.
\newblock In \emph{Proceedings of the IEEE/CVF International Conference on Computer Vision}, pages 2289--2298, 2021.

\bibitem[Arbelaez et~al.(2010)Arbelaez, Maire, Fowlkes, and Malik]{arbelaez2010contour}
Pablo Arbelaez, Michael Maire, Charless Fowlkes, and Jitendra Malik.
\newblock Contour detection and hierarchical image segmentation.
\newblock \emph{IEEE transactions on pattern analysis and machine intelligence}, 33\penalty0 (5):\penalty0 898--916, 2010.

\bibitem[Bevilacqua et~al.(2012)Bevilacqua, Roumy, Guillemot, and Alberi-Morel]{bevilacqua2012low}
Marco Bevilacqua, Aline Roumy, Christine Guillemot, and Marie~Line Alberi-Morel.
\newblock Low-complexity single-image super-resolution based on nonnegative neighbor embedding.
\newblock 2012.

\bibitem[Chen et~al.(2021)Chen, Wang, Guo, Xu, Deng, Liu, Ma, Xu, Xu, and Gao]{chen2021pre}
Hanting Chen, Yunhe Wang, Tianyu Guo, Chang Xu, Yiping Deng, Zhenhua Liu, Siwei Ma, Chunjing Xu, Chao Xu, and Wen Gao.
\newblock Pre-trained image processing transformer.
\newblock In \emph{Proceedings of the IEEE/CVF conference on computer vision and pattern recognition}, pages 12299--12310, 2021.

\bibitem[Chen et~al.(2023{\natexlab{a}})Chen, Wang, Zhang, Kong, Qiao, Zhou, and Dong]{chen2023hat}
Xiangyu Chen, Xintao Wang, Wenlong Zhang, Xiangtao Kong, Yu Qiao, Jiantao Zhou, and Chao Dong.
\newblock Hat: Hybrid attention transformer for image restoration.
\newblock \emph{arXiv preprint arXiv:2309.05239}, 2023{\natexlab{a}}.

\bibitem[Chen et~al.(2023{\natexlab{b}})Chen, Wang, Zhou, Qiao, and Dong]{chen2023activating}
Xiangyu Chen, Xintao Wang, Jiantao Zhou, Yu Qiao, and Chao Dong.
\newblock Activating more pixels in image super-resolution transformer.
\newblock In \emph{Proceedings of the IEEE/CVF Conference on Computer Vision and Pattern Recognition}, pages 22367--22377, 2023{\natexlab{b}}.

\bibitem[Chen et~al.(2022)Chen, Zhang, Gu, Zhang, Kong, and Yuan]{chen2022cross}
Zheng Chen, Yulun Zhang, Jinjin Gu, Yongbing Zhang, Linghe Kong, and Xin Yuan.
\newblock Cross aggregation transformer for image restoration.
\newblock In \emph{NeurIPS}, 2022.

\bibitem[Chen et~al.(2023{\natexlab{c}})Chen, Zhang, Gu, Kong, Yang, and Yu]{chen2023dual}
Zheng Chen, Yulun Zhang, Jinjin Gu, Linghe Kong, Xiaokang Yang, and Fisher Yu.
\newblock Dual aggregation transformer for image super-resolution.
\newblock In \emph{Proceedings of the IEEE/CVF international conference on computer vision}, pages 12312--12321, 2023{\natexlab{c}}.

\bibitem[Chu et~al.(2021)Chu, Tian, Wang, Zhang, Ren, Wei, Xia, and Shen]{Chu_2021_twin}
Xiangxiang Chu, Zhi Tian, Yuqing Wang, Bo Zhang, Haibing Ren, Xiaolin Wei, Huaxia Xia, and Chunhua Shen.
\newblock Twins: Revisiting the design of spatial attention in vision transformers, 2021.

\bibitem[Dai et~al.(2019)Dai, Cai, Zhang, Xia, and Zhang]{dai2019second}
Tao Dai, Jianrui Cai, Yongbing Zhang, Shu-Tao Xia, and Lei Zhang.
\newblock Second-order attention network for single image super-resolution.
\newblock In \emph{Proceedings of the IEEE/CVF conference on computer vision and pattern recognition}, pages 11065--11074, 2019.

\bibitem[Dai et~al.(2020)Dai, Cai, Zhang, Xia, and Zhang]{Dai_2020_san}
Tao Dai, Jianrui Cai, Yongbing Zhang, Shu-Tao Xia, and Lei Zhang.
\newblock Second-order attention network for single image super-resolution.
\newblock In \emph{2019 IEEE/CVF Conference on Computer Vision and Pattern Recognition (CVPR)}, 2020.

\bibitem[Dao and Gu(2024)]{mamba2}
Tri Dao and Albert Gu.
\newblock Transformers are {SSM}s: Generalized models and efficient algorithms through structured state space duality.
\newblock In \emph{International Conference on Machine Learning (ICML)}, 2024.

\bibitem[Ding et~al.(2022)Ding, Zhang, Han, and Ding]{ding2022scaling}
Xiaohan Ding, Xiangyu Zhang, Jungong Han, and Guiguang Ding.
\newblock Scaling up your kernels to 31x31: Revisiting large kernel design in cnns.
\newblock In \emph{Proceedings of the IEEE/CVF conference on computer vision and pattern recognition}, pages 11963--11975, 2022.

\bibitem[Dong et~al.(2014)Dong, Loy, He, and Tang]{dong2014learning}
Chao Dong, Chen~Change Loy, Kaiming He, and Xiaoou Tang.
\newblock Learning a deep convolutional network for image super-resolution.
\newblock In \emph{Computer Vision--ECCV 2014: 13th European Conference, Zurich, Switzerland, September 6-12, 2014, Proceedings, Part IV 13}, pages 184--199. Springer, 2014.

\bibitem[Dosovitskiy(2020)]{dosovitskiy2020image}
Alexey Dosovitskiy.
\newblock An image is worth 16x16 words: Transformers for image recognition at scale.
\newblock \emph{arXiv preprint arXiv:2010.11929}, 2020.

\bibitem[Franzen(2021)]{kao}
Rich Franzen.
\newblock Kodak lossless true color image suite.
\newblock 2021.

\bibitem[Gu and Dao(2023)]{mamba}
Albert Gu and Tri Dao.
\newblock Mamba: Linear-time sequence modeling with selective state spaces.
\newblock \emph{arXiv preprint arXiv:2312.00752}, 2023.

\bibitem[Guo et~al.(2025)Guo, Li, Dai, Ouyang, Ren, and Xia]{guo2025mambair}
Hang Guo, Jinmin Li, Tao Dai, Zhihao Ouyang, Xudong Ren, and Shu-Tao Xia.
\newblock Mambair: A simple baseline for image restoration with state-space model.
\newblock In \emph{European Conference on Computer Vision}, pages 222--241. Springer, 2025.

\bibitem[Huang et~al.(2015)Huang, Singh, and Ahuja]{huang2015single}
Jia-Bin Huang, Abhishek Singh, and Narendra Ahuja.
\newblock Single image super-resolution from transformed self-exemplars.
\newblock In \emph{Proceedings of the IEEE conference on computer vision and pattern recognition}, pages 5197--5206, 2015.

\bibitem[Karaali and Jung(2017)]{karaali2017edge}
Ali Karaali and Claudio~Rosito Jung.
\newblock Edge-based defocus blur estimation with adaptive scale selection.
\newblock \emph{IEEE Transactions on Image Processing}, 27\penalty0 (3):\penalty0 1126--1137, 2017.

\bibitem[Kim et~al.(2016{\natexlab{a}})Kim, Lee, and Lee]{Kim_2016_vdsr}
Jiwon Kim, Jung~Kwon Lee, and Kyoung~Mu Lee.
\newblock Accurate image super-resolution using very deep convolutional networks.
\newblock In \emph{2016 IEEE Conference on Computer Vision and Pattern Recognition (CVPR)}, 2016{\natexlab{a}}.

\bibitem[Kim et~al.(2016{\natexlab{b}})Kim, Lee, and Lee]{kim2016deeply}
Jiwon Kim, Jung~Kwon Lee, and Kyoung~Mu Lee.
\newblock Deeply-recursive convolutional network for image super-resolution.
\newblock In \emph{Proceedings of the IEEE conference on computer vision and pattern recognition}, pages 1637--1645, 2016{\natexlab{b}}.

\bibitem[Kingma and Ba(2014)]{kingma2014adam}
Diederik~P Kingma and Jimmy Ba.
\newblock Adam: A method for stochastic optimization.
\newblock \emph{arXiv preprint arXiv:1412.6980}, 2014.

\bibitem[Lee et~al.(2019)Lee, Lee, Cho, and Lee]{lee2019deep}
Junyong Lee, Sungkil Lee, Sunghyun Cho, and Seungyong Lee.
\newblock Deep defocus map estimation using domain adaptation.
\newblock In \emph{Proceedings of the IEEE/CVF conference on computer vision and pattern recognition}, pages 12222--12230, 2019.

\bibitem[Lee et~al.(2021)Lee, Son, Rim, Cho, and Lee]{lee2021iterative}
Junyong Lee, Hyeongseok Son, Jaesung Rim, Sunghyun Cho, and Seungyong Lee.
\newblock Iterative filter adaptive network for single image defocus deblurring.
\newblock In \emph{Proceedings of the IEEE/CVF Conference on Computer Vision and Pattern Recognition}, pages 2034--2042, 2021.

\bibitem[Li et~al.(2023{\natexlab{a}})Li, Fan, Xiang, Demandolx, Ranjan, Timofte, and Van~Gool]{li2023efficient}
Yawei Li, Yuchen Fan, Xiaoyu Xiang, Denis Demandolx, Rakesh Ranjan, Radu Timofte, and Luc Van~Gool.
\newblock Efficient and explicit modelling of image hierarchies for image restoration.
\newblock In \emph{Proceedings of the IEEE/CVF Conference on Computer Vision and Pattern Recognition}, pages 18278--18289, 2023{\natexlab{a}}.

\bibitem[Li et~al.(2023{\natexlab{b}})Li, Fan, Xiang, Demandolx, Ranjan, Timofte, and Van~Gool]{li2023grl}
Yawei Li, Yuchen Fan, Xiaoyu Xiang, Denis Demandolx, Rakesh Ranjan, Radu Timofte, and Luc Van~Gool.
\newblock Efficient and explicit modelling of image hierarchies for image restoration.
\newblock In \emph{Proceedings of the IEEE/CVF Conference on Computer Vision and Pattern Recognition}, pages 18278--18289, 2023{\natexlab{b}}.

\bibitem[Liang et~al.(2021)Liang, Cao, Sun, Zhang, Van~Gool, and Timofte]{liang2021swinir}
Jingyun Liang, Jiezhang Cao, Guolei Sun, Kai Zhang, Luc Van~Gool, and Radu Timofte.
\newblock Swinir: Image restoration using swin transformer.
\newblock In \emph{Proceedings of the IEEE/CVF international conference on computer vision}, pages 1833--1844, 2021.

\bibitem[Lim et~al.(2017{\natexlab{a}})Lim, Son, Kim, Nah, and Lee]{lim2017edsr}
Bee Lim, Sanghyun Son, Heewon Kim, Seungjun Nah, and Kyoung~Mu Lee.
\newblock Enhanced deep residual networks for single image super-resolution.
\newblock In \emph{2017 IEEE Conference on Computer Vision and Pattern Recognition Workshops (CVPRW)}, 2017{\natexlab{a}}.

\bibitem[Lim et~al.(2017{\natexlab{b}})Lim, Son, Kim, Nah, and Mu~Lee]{lim2017enhanced}
Bee Lim, Sanghyun Son, Heewon Kim, Seungjun Nah, and Kyoung Mu~Lee.
\newblock Enhanced deep residual networks for single image super-resolution.
\newblock In \emph{Proceedings of the IEEE conference on computer vision and pattern recognition workshops}, pages 136--144, 2017{\natexlab{b}}.

\bibitem[Liu et~al.(2024)Liu, Tian, Zhao, Yu, Xie, Wang, Ye, and Liu]{liu2024vmamba}
Yue Liu, Yunjie Tian, Yuzhong Zhao, Hongtian Yu, Lingxi Xie, Yaowei Wang, Qixiang Ye, and Yunfan Liu.
\newblock Vmamba: Visual state space model.
\newblock \emph{arXiv preprint arXiv:2401.10166}, 2024.

\bibitem[Liu et~al.(2021)Liu, Lin, Cao, Hu, Wei, Zhang, Lin, and Guo]{liu2021swin}
Ze Liu, Yutong Lin, Yue Cao, Han Hu, Yixuan Wei, Zheng Zhang, Stephen Lin, and Baining Guo.
\newblock Swin transformer: Hierarchical vision transformer using shifted windows.
\newblock In \emph{Proceedings of the IEEE/CVF international conference on computer vision}, pages 10012--10022, 2021.

\bibitem[Luo et~al.(2016)Luo, Li, Urtasun, and Zemel]{luo2016understanding}
Wenjie Luo, Yujia Li, Raquel Urtasun, and Richard Zemel.
\newblock Understanding the effective receptive field in deep convolutional neural networks.
\newblock \emph{Advances in neural information processing systems}, 29, 2016.

\bibitem[Ma et~al.(2016)Ma, Duanmu, Wu, Wang, Yong, Li, and Zhang]{ma2016waterloo}
Kede Ma, Zhengfang Duanmu, Qingbo Wu, Zhou Wang, Hongwei Yong, Hongliang Li, and Lei Zhang.
\newblock Waterloo exploration database: New challenges for image quality assessment models.
\newblock \emph{IEEE Transactions on Image Processing}, 26\penalty0 (2):\penalty0 1004--1016, 2016.

\bibitem[Martin et~al.(2001)Martin, Fowlkes, Tal, and Malik]{martin2001database}
David Martin, Charless Fowlkes, Doron Tal, and Jitendra Malik.
\newblock A database of human segmented natural images and its application to evaluating segmentation algorithms and measuring ecological statistics.
\newblock In \emph{Proceedings Eighth IEEE International Conference on Computer Vision. ICCV 2001}, pages 416--423. IEEE, 2001.

\bibitem[Matsui et~al.(2017)Matsui, Ito, Aramaki, Fujimoto, Ogawa, Yamasaki, and Aizawa]{matsui2017sketch}
Yusuke Matsui, Kota Ito, Yuji Aramaki, Azuma Fujimoto, Toru Ogawa, Toshihiko Yamasaki, and Kiyoharu Aizawa.
\newblock Sketch-based manga retrieval using manga109 dataset.
\newblock \emph{Multimedia Tools and Applications}, 76:\penalty0 21811--21838, 2017.

\bibitem[Mei et~al.(2020)Mei, Fan, Zhou, Huang, Huang, and Shi]{mei2020image}
Yiqun Mei, Yuchen Fan, Yuqian Zhou, Lichao Huang, Thomas~S Huang, and Honghui Shi.
\newblock Image super-resolution with cross-scale non-local attention and exhaustive self-exemplars mining.
\newblock In \emph{Proceedings of the IEEE/CVF conference on computer vision and pattern recognition}, pages 5690--5699, 2020.

\bibitem[Mei et~al.(2021{\natexlab{a}})Mei, Fan, and Zhou]{Mei_2021_nlsa}
Yiqun Mei, Yuchen Fan, and Yuqian Zhou.
\newblock Image super-resolution with non-local sparse attention.
\newblock In \emph{2021 IEEE/CVF Conference on Computer Vision and Pattern Recognition (CVPR)}, 2021{\natexlab{a}}.

\bibitem[Mei et~al.(2021{\natexlab{b}})Mei, Fan, and Zhou]{mei2021image}
Yiqun Mei, Yuchen Fan, and Yuqian Zhou.
\newblock Image super-resolution with non-local sparse attention.
\newblock In \emph{Proceedings of the IEEE/CVF Conference on Computer Vision and Pattern Recognition}, pages 3517--3526, 2021{\natexlab{b}}.

\bibitem[Niu et~al.(2020{\natexlab{a}})Niu, Wen, Ren, Zhang, Yang, Wang, Zhang, Cao, and Shen]{Niu_2020_han}
Ben Niu, Weilei Wen, Wenqi Ren, Xiangde Zhang, Lianping Yang, Shuzhen Wang, Kaihao Zhang, Xiaochun Cao, and Haifeng Shen.
\newblock \emph{Single Image Super-Resolution via a Holistic Attention Network}, page 191–207.
\newblock 2020{\natexlab{a}}.

\bibitem[Niu et~al.(2020{\natexlab{b}})Niu, Wen, Ren, Zhang, Yang, Wang, Zhang, Cao, and Shen]{niu2020single}
Ben Niu, Weilei Wen, Wenqi Ren, Xiangde Zhang, Lianping Yang, Shuzhen Wang, Kaihao Zhang, Xiaochun Cao, and Haifeng Shen.
\newblock Single image super-resolution via a holistic attention network.
\newblock In \emph{Computer Vision--ECCV 2020: 16th European Conference, Glasgow, UK, August 23--28, 2020, Proceedings, Part XII 16}, pages 191--207. Springer, 2020{\natexlab{b}}.

\bibitem[Plotz and Roth(2017)]{plotz2017benchmarking}
Tobias Plotz and Stefan Roth.
\newblock Benchmarking denoising algorithms with real photographs.
\newblock In \emph{Proceedings of the IEEE conference on computer vision and pattern recognition}, pages 1586--1595, 2017.

\bibitem[Ray et~al.(2024)Ray, Kumar, and Kolekar]{ray2024cfat}
Abhisek Ray, Gaurav Kumar, and Maheshkumar~H Kolekar.
\newblock Cfat: Unleashing triangularwindows for image super-resolution.
\newblock \emph{arXiv preprint arXiv:2403.16143}, 2024.

\bibitem[Ren et~al.(2021)Ren, He, Wang, and Zhao]{ren2021adaptive}
Chao Ren, Xiaohai He, Chuncheng Wang, and Zhibo Zhao.
\newblock Adaptive consistency prior based deep network for image denoising.
\newblock In \emph{Proceedings of the IEEE/CVF conference on computer vision and pattern recognition}, pages 8596--8606, 2021.

\bibitem[Shi et~al.(2015)Shi, Xu, and Jia]{shi2015just}
Jianping Shi, Li Xu, and Jiaya Jia.
\newblock Just noticeable defocus blur detection and estimation.
\newblock In \emph{Proceedings of the IEEE Conference on Computer Vision and Pattern Recognition}, pages 657--665, 2015.

\bibitem[Son et~al.(2021)Son, Lee, Cho, and Lee]{son2021single}
Hyeongseok Son, Junyong Lee, Sunghyun Cho, and Seungyong Lee.
\newblock Single image defocus deblurring using kernel-sharing parallel atrous convolutions.
\newblock In \emph{Proceedings of the IEEE/CVF International Conference on Computer Vision}, pages 2642--2650, 2021.

\bibitem[Timofte et~al.(2017)Timofte, Agustsson, Van~Gool, Yang, and Zhang]{timofte2017ntire}
Radu Timofte, Eirikur Agustsson, Luc Van~Gool, Ming-Hsuan Yang, and Lei Zhang.
\newblock Ntire 2017 challenge on single image super-resolution: Methods and results.
\newblock In \emph{Proceedings of the IEEE conference on computer vision and pattern recognition workshops}, pages 114--125, 2017.

\bibitem[Waleffe et~al.(2024)Waleffe, Byeon, Riach, Norick, Korthikanti, Dao, Gu, Hatamizadeh, Singh, Narayanan, et~al.]{waleffe2024empirical}
Roger Waleffe, Wonmin Byeon, Duncan Riach, Brandon Norick, Vijay Korthikanti, Tri Dao, Albert Gu, Ali Hatamizadeh, Sudhakar Singh, Deepak Narayanan, et~al.
\newblock An empirical study of mamba-based language models.
\newblock \emph{arXiv preprint arXiv:2406.07887}, 2024.

\bibitem[Wang et~al.(2023)Wang, Chen, Ni, Liu, and jinfan]{omni_sr}
Hang Wang, Xuanhong Chen, Bingbing Ni, Yutian Liu, and Liu jinfan.
\newblock Omni aggregation networks for lightweight image super-resolution.
\newblock In \emph{Conference on Computer Vision and Pattern Recognition}, 2023.

\bibitem[Wang et~al.(2022{\natexlab{a}})Wang, Xie, Li, Fan, Song, Liang, Lu, Luo, and Shao]{Wang_2022_pvt}
Wenhai Wang, Enze Xie, Xiang Li, Deng-Ping Fan, Kaitao Song, Ding Liang, Tong Lu, Ping Luo, and Ling Shao.
\newblock Pyramid vision transformer: A versatile backbone for dense prediction without convolutions.
\newblock In \emph{2021 IEEE/CVF International Conference on Computer Vision (ICCV)}, 2022{\natexlab{a}}.

\bibitem[Wang et~al.(2022{\natexlab{b}})Wang, Cun, Bao, Zhou, Liu, and Li]{wang2022uformer}
Zhendong Wang, Xiaodong Cun, Jianmin Bao, Wengang Zhou, Jianzhuang Liu, and Houqiang Li.
\newblock Uformer: A general u-shaped transformer for image restoration.
\newblock In \emph{Proceedings of the IEEE/CVF Conference on Computer Vision and Pattern Recognition (CVPR)}, pages 17683--17693, 2022{\natexlab{b}}.

\bibitem[Wen et~al.(2023)Wen, Cheng, Xu, Zhou, Timofte, Hou, and Van~Gool]{wen2023super}
Juan Wen, Shupeng Cheng, Peng Xu, Bowen Zhou, Radu Timofte, Weiyan Hou, and Luc Van~Gool.
\newblock When super-resolution meets camouflaged object detection: A comparison study.
\newblock \emph{arXiv preprint arXiv:2308.04370}, 2023.

\bibitem[Wen et~al.(2024)Wen, Li, Zhang, Hou, Timofte, and Van~Gool]{wen2024empowering}
Juan Wen, Yawei Li, Chao Zhang, Weiyan Hou, Radu Timofte, and Luc Van~Gool.
\newblock Empowering image recovery\_ a multi-attention approach.
\newblock \emph{arXiv preprint arXiv:2404.04617}, 2024.

\bibitem[Zamir et~al.(2021)Zamir, Arora, Khan, Hayat, Khan, Yang, and Shao]{zamir2021multi}
Syed~Waqas Zamir, Aditya Arora, Salman Khan, Munawar Hayat, Fahad~Shahbaz Khan, Ming-Hsuan Yang, and Ling Shao.
\newblock Multi-stage progressive image restoration.
\newblock In \emph{Proceedings of the IEEE/CVF conference on computer vision and pattern recognition}, pages 14821--14831, 2021.

\bibitem[Zamir et~al.(2022)Zamir, Arora, Khan, Hayat, Khan, and Yang]{zamir2022restormer}
Syed~Waqas Zamir, Aditya Arora, Salman Khan, Munawar Hayat, Fahad~Shahbaz Khan, and Ming-Hsuan Yang.
\newblock Restormer: Efficient transformer for high-resolution image restoration.
\newblock In \emph{Proceedings of the IEEE/CVF conference on computer vision and pattern recognition}, pages 5728--5739, 2022.

\bibitem[Zeyde et~al.(2012)Zeyde, Elad, and Protter]{zeyde2012single}
Roman Zeyde, Michael Elad, and Matan Protter.
\newblock On single image scale-up using sparse-representations.
\newblock In \emph{Curves and Surfaces: 7th International Conference, Avignon, France, June 24-30, 2010, Revised Selected Papers 7}, pages 711--730. Springer, 2012.

\bibitem[Zhang et~al.(2023)Zhang, Zhang, Gu, Zhang, Kong, and Yuan]{zhang2023accurate}
Jiale Zhang, Yulun Zhang, Jinjin Gu, Yongbing Zhang, Linghe Kong, and Xin Yuan.
\newblock Accurate image restoration with attention retractable transformer.
\newblock In \emph{ICLR}, 2023.

\bibitem[Zhang et~al.(2017{\natexlab{a}})Zhang, Zuo, Chen, Meng, and Zhang]{DnCNN}
Kai Zhang, Wangmeng Zuo, Yunjin Chen, Deyu Meng, and Lei Zhang.
\newblock Beyond a gaussian denoiser: Residual learning of deep cnn for image denoising.
\newblock \emph{IEEE transactions on image processing}, 26\penalty0 (7):\penalty0 3142--3155, 2017{\natexlab{a}}.

\bibitem[Zhang et~al.(2017{\natexlab{b}})Zhang, Zuo, Gu, and Zhang]{IRCNN}
Kai Zhang, Wangmeng Zuo, Shuhang Gu, and Lei Zhang.
\newblock Learning deep cnn denoiser prior for image restoration.
\newblock In \emph{Proceedings of the IEEE conference on computer vision and pattern recognition}, pages 3929--3938, 2017{\natexlab{b}}.

\bibitem[Zhang et~al.(2018{\natexlab{a}})Zhang, Zuo, and Zhang]{FFDNet}
Kai Zhang, Wangmeng Zuo, and Lei Zhang.
\newblock Ffdnet: Toward a fast and flexible solution for cnn-based image denoising.
\newblock \emph{IEEE Transactions on Image Processing}, 27\penalty0 (9):\penalty0 4608--4622, 2018{\natexlab{a}}.

\bibitem[Zhang et~al.(2021)Zhang, Li, Zuo, Zhang, Van~Gool, and Timofte]{DRUNet}
Kai Zhang, Yawei Li, Wangmeng Zuo, Lei Zhang, Luc Van~Gool, and Radu Timofte.
\newblock Plug-and-play image restoration with deep denoiser prior.
\newblock \emph{IEEE Transactions on Pattern Analysis and Machine Intelligence}, 44\penalty0 (10):\penalty0 6360--6376, 2021.

\bibitem[Zhang et~al.(2011)Zhang, Wu, Buades, and Li]{zhang2011color}
Lei Zhang, Xiaolin Wu, Antoni Buades, and Xin Li.
\newblock Color demosaicking by local directional interpolation and nonlocal adaptive thresholding.
\newblock \emph{Journal of Electronic imaging}, 20\penalty0 (2):\penalty0 023016--023016, 2011.

\bibitem[Zhang et~al.(2024)Zhang, Li, Zhou, Zhao, and Gu]{zhang2024transcending}
Leheng Zhang, Yawei Li, Xingyu Zhou, Xiaorui Zhao, and Shuhang Gu.
\newblock Transcending the limit of local window: Advanced super-resolution transformer with adaptive token dictionary.
\newblock In \emph{Proceedings of the IEEE/CVF Conference on Computer Vision and Pattern Recognition}, pages 2856--2865, 2024.

\bibitem[Zhang et~al.(2022)Zhang, Zeng, Guo, and Zhang]{zhang2022efficient}
Xindong Zhang, Hui Zeng, Shi Guo, and Lei Zhang.
\newblock Efficient long-range attention network for image super-resolution.
\newblock In \emph{European Conference on Computer Vision}, pages 649--667. Springer, 2022.

\bibitem[Zhang et~al.(2018{\natexlab{b}})Zhang, Li, Li, Wang, Zhong, and Fu]{zhang2018image}
Yulun Zhang, Kunpeng Li, Kai Li, Lichen Wang, Bineng Zhong, and Yun Fu.
\newblock Image super-resolution using very deep residual channel attention networks.
\newblock In \emph{Proceedings of the European conference on computer vision (ECCV)}, pages 286--301, 2018{\natexlab{b}}.

\bibitem[Zhang et~al.(2018{\natexlab{c}})Zhang, Li, Li, Wang, Zhong, and Fu]{zhang2018rcan}
Yulun Zhang, Kunpeng Li, Kai Li, Lichen Wang, Bineng Zhong, and Yun Fu.
\newblock \emph{Image Super-Resolution Using Very Deep Residual Channel Attention Networks}, page 294–310.
\newblock 2018{\natexlab{c}}.

\bibitem[Zhang et~al.(2018{\natexlab{d}})Zhang, Tian, Kong, Zhong, and Fu]{Zhang_2018_rdn}
Yulun Zhang, Yapeng Tian, Yu Kong, Bineng Zhong, and Yun Fu.
\newblock Residual dense network for image super-resolution, 2018{\natexlab{d}}.

\bibitem[Zhang et~al.(2018{\natexlab{e}})Zhang, Tian, Kong, Zhong, and Fu]{zhang2018residual}
Yulun Zhang, Yapeng Tian, Yu Kong, Bineng Zhong, and Yun Fu.
\newblock Residual dense network for image super-resolution.
\newblock In \emph{Proceedings of the IEEE conference on computer vision and pattern recognition}, pages 2472--2481, 2018{\natexlab{e}}.

\bibitem[Zhou et~al.(2020)Zhou, Zhang, Zuo, and Loy]{zhou2020cross}
Shangchen Zhou, Jiawei Zhang, Wangmeng Zuo, and Chen~Change Loy.
\newblock Cross-scale internal graph neural network for image super-resolution.
\newblock In \emph{Advances in Neural Information Processing Systems}, 2020.

\bibitem[Zhou et~al.(2023)Zhou, Li, Guo, Bai, Cheng, and Hou]{zhou2023srformer}
Yupeng Zhou, Zhen Li, Chun-Le Guo, Song Bai, Ming-Ming Cheng, and Qibin Hou.
\newblock Srformer: Permuted self-attention for single image super-resolution.
\newblock \emph{arXiv preprint arXiv:2303.09735}, 2023.

\end{thebibliography}
